\documentclass[letterpaper]{article}
\usepackage[preprint]{aaai2027}

\usepackage[hyphens]{url}
\usepackage{graphicx}
\usepackage{natbib}
\usepackage{caption}
\usepackage{algorithm}
\usepackage{algorithmic}
\usepackage{amsmath}
\usepackage{amssymb}
\usepackage{booktabs}
\usepackage{multirow}
\usepackage[capitalize,noabbrev]{cleveref}

\title{$P^3$: Probabilistic Policy Propagation for Stable VAE-Based Robot Learning}
\author{
    Liyun Yan\textsuperscript{\rm 1,4},
    Jianming Ma\textsuperscript{\rm 1,4},
    Yang Zhang\textsuperscript{\rm 1},\\
    Shengcheng Fu\textsuperscript{\rm 2},
    Zhanxiang Cao\textsuperscript{\rm 1,4},
    Keqi Zhu\textsuperscript{\rm 3,4},\\
    Yizhi Chen\textsuperscript{\rm 2,4},
    Yue Gao\textsuperscript{\rm 1,4}\corresponding
}
\affiliations{
    \textsuperscript{\rm 1}Shanghai Jiao Tong University, Shanghai, China\\
    \textsuperscript{\rm 2}Tongji University, Shanghai, China\\
    \textsuperscript{\rm 3}Zhejiang University, Hangzhou, China\\
    \textsuperscript{\rm 4}Shanghai Innovation Institute, Shanghai, China\\
    yuegao@sjtu.edu.cn
}

\begin{document}

\maketitle

\begin{abstract}
Variational Autoencoders are widely used to encode high-dimensional and noisy observations in robotics. 
However, their stochastic latent creates a mismatch with Proximal Policy Optimization (PPO): an effective policy marginalizes over the latent distribution, whereas former implementations estimate its probability ratio and KL divergence using only one latent sample.
We identify a fundamental but overlooked theoretical cause: naive single-sample approximations in stochastic latent space induce significant variance and bias in the surrogate loss. 
To address this, we introduce $P^3$ (Probabilistic Policy Propagation), a distribution-aware optimization framework for VAE-based policies. $P^3$ couples moment-based probabilistic method for stable and efficient learning with sampling-based calibration for robust policy behavior under latent uncertainty.
In our experiments, $P^3$ boosts data efficiency from $64.6\%$ to $>96\%$, reduces convergence steps by $>20\%$.
Furthermore, $P^3$ is evaluated on challenging humanoid parkour tasks and shows an effective foundation for VAE-based PPO.
Code is available at \url{https://github.com/ylyem9x/P3_Open}.

\end{abstract}

\section{Introduction}
\label{sec:introduction}

The integration of Variational Autoencoder (VAE)~\cite{vae,betaVAE} with Proximal Policy Optimization (PPO)~\cite{ppo} has become a widely adopted framework in learning-based legged locomotion.
In this framework, the VAE compresses high-dimensional proprioceptive and exteroceptive data into a compact latent representation~\cite{understanding,denoising}, which serves as an informative observation for the actor.
By coupling state estimation with control, this framework has supported remarkable progress in complex-terrain locomotion and robust sim-to-real transfer for both quadrupedal~\cite{pie, terrain_recon, move, dreamwaq2, moral, dreamflex, lip, mbc} and humanoid~\cite{PIM, czx, ddl, ddlv, Noetix}.

The widespread adoption of this framework stems from its ability to address real-world partial observability~\cite{concurrent,rma,stateesimation}, where high-dimensional, noisy observation histories must be distilled into task-relevant latent states.
Moreover, by training on stochastic samples from this latent distribution, the framework forces the policy to adapt to representation uncertainty and filter out sensor noise~\cite{him}.
These properties make VAE-based PPO an effective framework for learning robust robot-control policies.

Despite its empirical success, the stochastic latent representation introduces an optimization issue that standard PPO does not explicitly account for.
Existing implementations, even those employing heuristics such as adaptive bootstrapping~\cite{dreamwaq}, often exhibit slow convergence, training instability, and suboptimal asymptotic performance~\cite{pvp}.
This raises a key question: why is this effective framework difficult to optimize reliably?

\begin{figure}[t]
  \centering
  \includegraphics[width=\columnwidth]{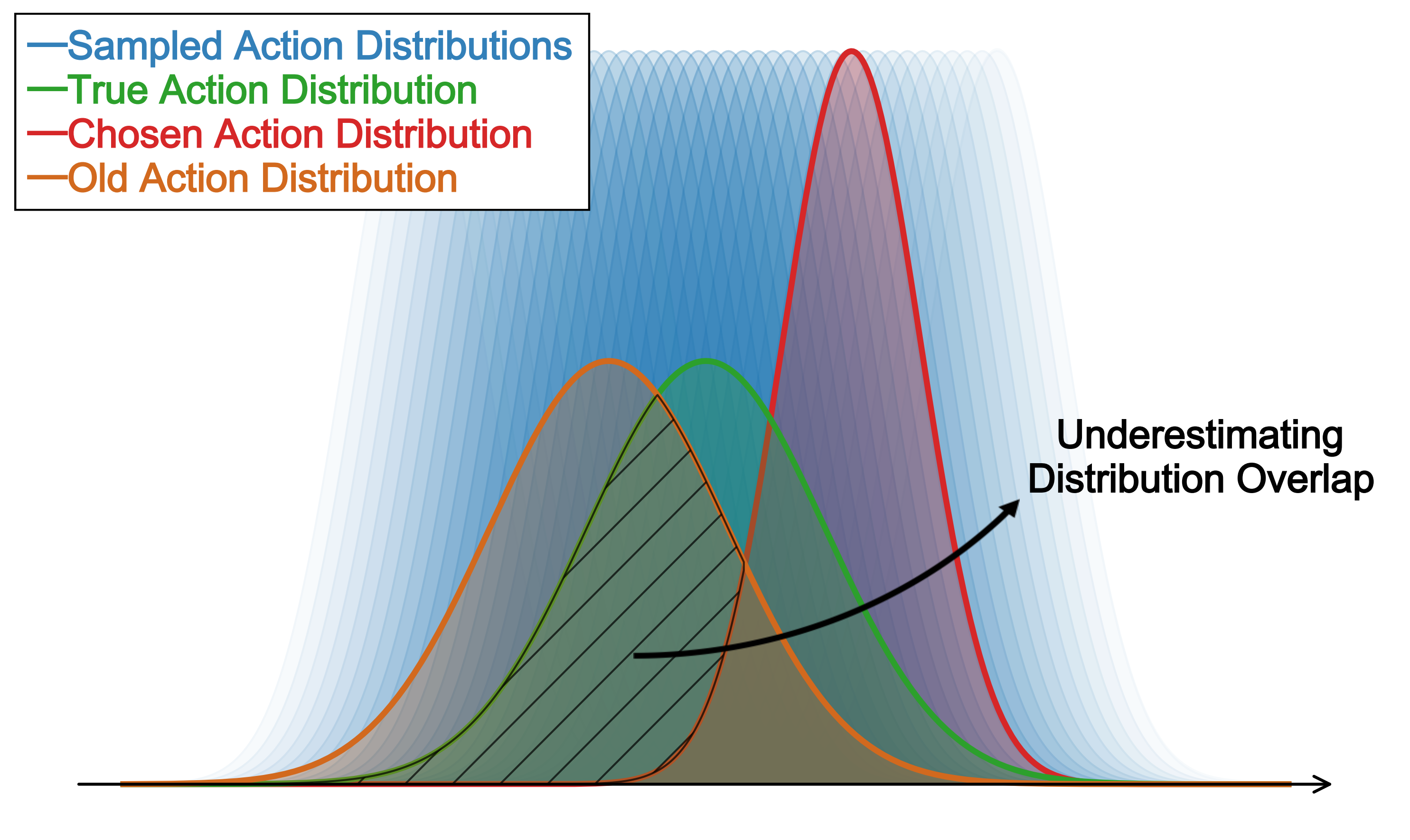}
  \caption{
    \textbf{Single-sample likelihood mismatch in VAE-based PPO.}
    Each latent sample induces an action component (blue), while the marginalized policy aggregates them (green).
    Using one component (red) underestimates its overlap with the old policy (orange), distorting the ratio and KL divergence used by PPO.
  }
  \label{fig:intro_pic}
\end{figure}

In this work, we identify a previously overlooked theoretical problem in VAE-based PPO.
Each observation defines a latent distribution, and each latent sample induces a different action distribution.
PPO should compare the aggregate distributions---the marginalized policies---whereas prior methods compare one sampled component from each policy.
As shown in Figure~\ref{fig:intro_pic}, this approximation biases KL estimation and increases the variance of surrogate-loss gradients, causing erroneous clipping and unstable optimization.

To address this problem, we introduce $P^3$ (Probabilistic Policy Propagation), a distribution-aware framework with complementary estimators of the marginalized policy (Figure~\ref{fig:main_pic}).
Deterministic moment matching (MM)~\cite{pbp,dvi,AnaCov,lightweight,mombo} enables stable, efficient learning, while a Monte Carlo (MC) latent-sampling refinement~\cite{pets} improves robustness to latent uncertainty.

Our main contributions are summarized as follows:
\begin{itemize}
    \item We explain why prior methods exhibit slow, unstable training and suboptimal asymptotic performance.
    \item We introduce $P^3$, which propagates the latent distribution through the actor to combine efficient optimization with accurate uncertainty estimation.
    \item Experiments show that $P^3$ raises data efficiency from $64.6\%$ to $>96\%$, reduces convergence steps by $>20\%$, and achieves the best transfer results.
\end{itemize}

\section{Preliminaries: VAE-based Legged Locomotion}
\label{sec:preliminaries}

\paragraph{POMDP Formulation.} Legged locomotion is standardly modeled as a Partially Observable Markov Decision Process (POMDP), defined by the tuple $\mathcal{M} = (\mathcal{S}, \mathcal{O}, \mathcal{A}, \mathcal{T}, \mathcal{R}, \gamma)$. 
At each timestep $t$, the agent receives a partial observation $o_t \in \mathcal{O}$ and executes an action $a_t \in \mathcal{A}$. 
The environment transitions according to $s_{t+1} \sim \mathcal{T}(s_{t+1} \mid s_t, a_t)$ and yields a reward $r_t$. 
The goal is to learn a policy that maximizes the expected cumulative discounted return $J(\pi) = \mathbb{E}_{\tau \sim \pi} [\sum_{t=0}^{\infty} \gamma^t r_t]$. 

\paragraph{VAE-based State Estimation.} Since critical states (e.g., linear velocity $v_t$ and terrain friction) are often unobservable, recent methods integrate Variational Autoencoders (VAEs) for state estimation. 
\cite{dreamwaq} pioneered the use of proprioceptive history $o^H_t$ to encode a latent variable $z_t$, reconstructing future observations to implicitly model environments and dynamics. 
Subsequent works like PIE~\cite{pie} incorporate exteroceptive data $o^{extero}$ into the encoder and reconstruction target $\hat{y}_t$ (e.g., height maps), enabling explicit modeling of complex terrain.

\paragraph{Policy Structure and Optimization.} This framework couples a stochastic encoder $q_\phi(z_t \mid o^H_t)$ with a policy network $p_\psi(a_t \mid z_t, o_t)$ (denoted as $q_\phi(z_t \mid o_t)$ and $p_\psi(a_t \mid z_t)$ hereinafter). 
Consequently, the effective control policy is formulated as a marginal distribution over the latent space:
\begin{equation}
    \label{eq:marginal_policy}
    \pi_\theta(a_t \mid o_t) = \int p_\psi(a_t \mid z_t) q_\phi(z_t \mid o_t) \, d z_t.
\end{equation}
End-to-end training of $\theta=(\phi,\psi)$ minimizes the combined clipped policy-gradient and VAE loss:
\begin{equation}
    \label{eq:total_loss}
    \begin{aligned}
        & \mathcal{L}_{total} = \mathcal{L}_{CLIP}(\theta) +  \mathcal{L}_{VAE}(\phi), \\
        & \mathcal{L}_{CLIP}(\theta) = \hat{\mathbb{E}} \left[ - \min( r_\theta \hat{A}, \text{clip}(r_\theta, 1-\epsilon, 1+\epsilon) \hat{A} ) \right], \\
        & \mathcal{L}_{VAE}(\phi) = MSE(y_t) + \beta D_{KL}(q_\phi(z_t \mid o_t) \| \mathcal{N}(0, I)). 
    \end{aligned} 
\end{equation}
Here, $r_\theta=\pi_\theta(a\mid o)/\pi_{\theta_{old}}(a\mid o)$ denotes the probability ratio, $\hat{A}$ is the advantage estimate, $MSE(y_t)$ represents the reconstruction error, and $\beta$ is the weighting coefficient.
This joint optimization aims to learn a robust latent representation while simultaneously maximizing task performance.

\section{Theoretical Analysis}
\label{sec:theory}
As illustrated in \cref{fig:intro_pic}, each latent sample $z \sim q_\phi(\cdot|o)$ induces a Gaussian distribution in the action space via the actor network $p_\psi(\cdot|z)$ and policy standard deviation $\sigma_{act}$ for PPO exploration. 
The overall policy $\pi_\theta(a|o)$, which marginalizes over the latent space, aggregates these components and forms a distribution with higher variance.

However, replacing the marginalized policy with a single sampled component creates two distinct estimation errors. 
First, it systematically distorts the log probability ratio, changing PPO's clipping decisions and yielding a biased KL estimate after averaging over actions. 
Second, finite latent sampling makes the probability-ratio estimate fluctuate across evaluations, increasing the variance of the surrogate-loss gradient and destabilizing optimization.

\subsection{KL Divergence Analysis}
To further elucidate how this systematic distortion undermines trust-region enforcement in PPO, we use the connection between the KL divergence and the probability ratio, $D_{KL}(\pi_{\theta_{old}} \parallel \pi_\theta) = \mathbb{E}_{a \sim \pi_{\theta_{old}}}[-\log r_\theta(a)]$, and first examine the decomposed policy KL divergence.
We substitute the marginalized policy $\pi_\theta(a_t \mid o_t) = \int p_\psi(a_t \mid z_t) q_\phi(z_t \mid o_t) \, d z_t$ into the KL divergence formula. 
By applying Jensen's inequality, we obtain a tractable upper bound for the policy KL divergence (detailed in Appendix~\ref{sec:kl_de}):
\begin{equation} 
    \label{eq:kl_decomposition} 
    \begin{aligned} 
        &D_{KL}(\pi_{\theta_{old}}(\cdot|o) \parallel \pi_\theta(\cdot|o)) \\ 
        &\le \mathbb{E}_{z \sim q_{\phi_{old}}} \Big[ D_{KL}\big(p_{\psi_{old}}(\cdot|z) \parallel p_\psi(\cdot|z)\big) \Big] \\
        & \quad + D_{KL}\big(q_{\phi_{old}} \parallel q_\phi\big). 
    \end{aligned} 
\end{equation}
To quantify the impact of this decomposition on optimization stability, we recall the theoretical monotonic improvement lower bound derived in TRPO~\cite{trpo}:
\begin{equation}
\label{eq:opt_bound}
    J(\pi_\theta) \ge L(\pi_\theta) - C \cdot D^{\max}_{KL}(\pi_{\theta_{old}} \parallel \pi_\theta),
\end{equation}
where $L(\pi_\theta)$ is the unclipped surrogate objective. 
While PPO approximates this trust region constraint via clipping, substituting our decomposition upper bound (\cref{eq:kl_decomposition}) into \cref{eq:opt_bound} yields this latent-conditioned-policy objective:
\begin{equation}
    \label{eq:opt_bound2}
    \begin{aligned}
        J(\pi_\theta) \ge & L(\pi_\theta) \\
        & -C \cdot\mathbb{E}^{max}_{z \sim q_{\phi_{old}}} \Big[ D_{KL}\big(p_{\psi_{old}}(\cdot|z) \parallel p_\psi(\cdot|z)\big) \Big] \\
        & - C \cdot D^{max}_{KL}\big(q_{\phi_{old}} \parallel q_\phi\big).
    \end{aligned}
\end{equation}
This derivation highlights that when optimizing a policy with a latent state estimator, the objective must simultaneously constrain the estimator's distribution shift ($D^{max}_{KL}(q_{\phi_{old}} \parallel q_\phi)$) and the actor's consistency across the entire latent distribution. 
While the former is implicitly regularized by the KL term in the VAE's loss function, prior methods often neglect the consistency term $\mathbb{E}^{max}_{z \sim q_{\phi_{old}}} \Big[ D_{KL}\big(p_{\psi_{old}}(\cdot|z) \parallel p_\psi(\cdot|z)\big) \Big]$, instead regularizing it only with respect to a single latent sample. 
In the PPO-clip objective, the component-to-marginal mismatch systematically distorts the log probability ratio.
These distorted probability ratios erroneously clip beneficial actions, while failing to penalize policy updates that exceed the trust region, leading to training inefficiency and instability.

\subsection{Gradient Noise Analysis}
We also perform an analysis of the additional noise ($\text{Noise}_{latent}$) introduced into the gradient by the stochastic sampling process. 
Our derivation (detailed in Appendices~\ref{sec:variance_ratio} and~\ref{sec:gradient_variance}) utilizes the Delta method and a Taylor expansion (conditions validated in \cref{subsec:action_analysis}) of the actor network to approximate the variance of the gradient $\hat{g}$.
The analysis reveals that the total variance of the gradient estimator constitutes the sum of the intrinsic PPO gradient variance and the additional noise induced by latent sampling.

The analytical expression for additional noise suggests:
\begin{equation}
    \begin{aligned}
    \text{Noise}_{latent} \propto \frac{A^2}{N} \cdot \underbrace{\left(\frac{d_a}{\sigma_{act}^2}\right)}_{\text{Score Var}} \cdot \underbrace{\left( \frac{\sigma_{vae}^2 \|\mathbf{J}_\psi\|_F^2}{\sigma_{act}^2} \right)}_{\text{Ratio Var}}.
    \end{aligned}
\end{equation}
$N$ is the number of latent samples, $d_a$ is the action dimension, $\sigma_{act}$ is the action standard deviation, and $\|\mathbf{J}_\psi\|_F$ is the Frobenius norm of the actor Jacobian.
The term $\text{Score Var}$ stems from the error in gradient calculation via PPO action sampling, while $\text{Ratio Var}$ arises from latent-space sampling.

From this equation, we draw the following conclusions regarding the optimization dynamics:

\paragraph{Adam-style optimizers} Adam's update step is inversely scaled by the estimated gradient to ensure stability~\cite{adam}. 
When facing the elevated gradient noise introduced by the latent sampling, the optimizer is forced to adopt aggressively smaller step sizes. 
More critically, as the policy attempts to converge—characterized by a reduction in $\sigma_{act}$—the gradient noise explodes. 
This surge in variance triggers an excessive reduction in the effective learning rate, causing the training to stagnate precisely when fine-grained policy refinement is required.

\paragraph{SGD-style optimizers} Characterized by stochastic gradient estimation, the optimization trajectory tends to naturally escape regions of high gradient variance due to the lack of adaptive moment estimation. 
SGD introduces an implicit bias that hinders the policy from reducing $\sigma_{act}$, making it difficult to converge towards a deterministic optimal policy. 

Additionally, the noise magnitude is directly correlated with the action dimension $d_a$. 
In high-degree-of-freedom tasks, such as humanoid control, this scaling effect further exacerbates the optimization difficulties.

Therefore, to retain the robustness of the VAE encoder while mitigating erroneous trust-region clipping and convergence stagnation, PPO should estimate the marginalized action likelihood rather than rely on a single latent sample.
This requires propagating the VAE latent distribution through the actor, motivating the probability estimators introduced next.

\section{Method}
\label{sec:p3}

\begin{figure*}[t]
  \centering
  \includegraphics[width=0.9\textwidth]{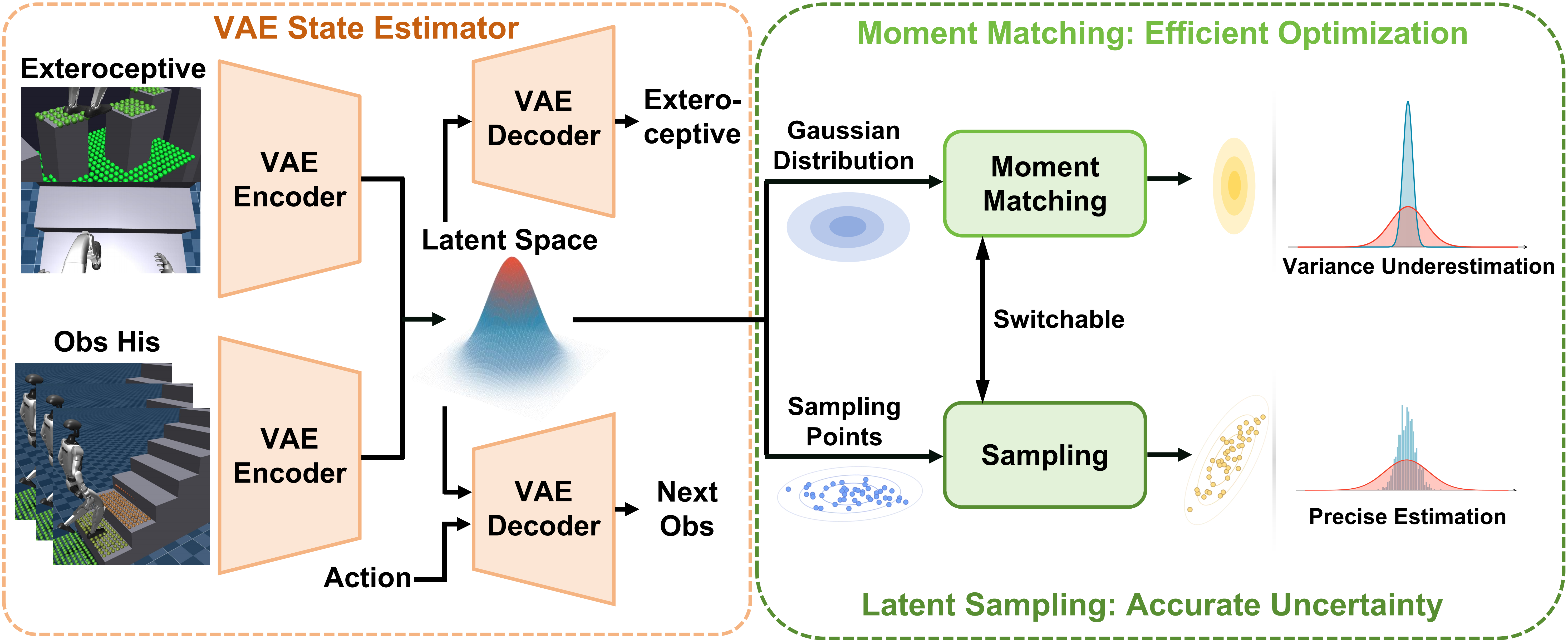}
  \caption{
    \textbf{Overview of $P^3$.}
    The left panel illustrates the VAE-based state estimator. The right panel shows two complementary estimators of the same marginalized policy: moment matching (MM) provides efficient, low-noise optimization, whereas Monte Carlo (MC) latent sampling more faithfully exposes the actor to latent uncertainty. Both operate on the same actor and can be switched without changing its architecture or parameters.
  }
  \label{fig:main_pic}
\end{figure*}

We propose Probabilistic Policy Propagation ($P^3$) to estimate the marginalized policy in \cref{eq:marginal_policy} for PPO.
$P^3$ provides two interchangeable estimators: deterministic moment matching (MM) and Monte Carlo (MC) latent sampling.
They share the same objective and checkpoints, differing only in how latent uncertainty is propagated to the action likelihood.
MM favors efficient, low-noise optimization, whereas MC more faithfully captures latent uncertainty at higher computational cost.
Our default instantiation combines their strengths by learning a strong policy with MM and then applying a short sampling-based calibration, termed \emph{Latent Sample Fine-Tuning} (LSFT).

\subsection{Moment-Matching Estimator}
\label{subsec:mm_estimator}

MM evaluates the actor by propagating the first two moments $(\boldsymbol{\mu},\mathbf{v})$ through its existing layers under a diagonal-covariance approximation.
This is a probabilistic evaluation rule for the same weights $\psi$, rather than a separate actor network.
For deterministic actor inputs such as the previous action, the input variance is zero; the VAE latent contributes $(\boldsymbol{\mu}_z,\boldsymbol{\sigma}_z^2)$.
The propagated action variance is combined with the actor's exploration variance to give
\begin{equation}
    \widehat{\pi}^{\mathrm{MM}}_\theta(a\mid o)
    = \mathcal{N}\!\left(a\,\middle|\,\boldsymbol{\mu}_{out},
    \operatorname{diag}(\mathbf{v}_{out})+\sigma_{act}^2\mathbf{I}\right).
    \label{eq:mm_policy}
\end{equation}

\paragraph{Probabilistic Linear Layer.}
Given input moments $\boldsymbol{\mu}_{in}$ and $\mathbf{v}_{in}$, the output moments for a linear layer $(\mathbf{W},\mathbf{b})$ are
\begin{equation}
\begin{aligned}
\boldsymbol{\mu}_{out} &= \mathbf{W}\boldsymbol{\mu}_{in} + \mathbf{b}, \\
\mathbf{v}_{out} &= (\mathbf{W} \circ \mathbf{W}) \mathbf{v}_{in},
\end{aligned}
\end{equation}
where $\circ$ is the Hadamard product.

\paragraph{Probabilistic ELU Activation.}
For an input $x\sim\mathcal{N}(\mu,\sigma^2)$, we derive the analytical moments of the ELU activation (Appendix~\ref{sec:elu_derivation}).
Let $\Phi$ and $\phi$ denote the CDF and PDF of the standard normal distribution, respectively, and let $\beta=\mu/\sigma$.
The expectation is
\begin{equation}
\mathbb{E}[y] = \mu \Phi(\beta) + \sigma \phi(\beta) + \alpha \left[ e^{\mu + \frac{\sigma^2}{2}} \Phi\left(-\beta - \sigma\right) - \Phi(-\beta) \right].
\end{equation}

The second moment is
\begin{equation}
\begin{split}
\mathbb{E}[y^2] &= (\mu^2 + \sigma^2)\Phi(\beta) + \mu\sigma\phi(\beta) \\
&\quad + \alpha^2 \Big[ e^{2\mu + 2\sigma^2} \Phi(-\beta - 2\sigma) \\ 
&\quad - 2 e^{\mu + \frac{\sigma^2}{2}} \Phi(-\beta - \sigma) + \Phi(-\beta) \Big].
\end{split}
\end{equation}

The output variance is $\mathbf{v}_{out} = \mathbb{E}[y^2] - (\mathbb{E}[y])^2$.

\paragraph{Optimization Properties.}
MM is deterministic conditional on $o_t$: repeated evaluations produce the same marginalized-likelihood approximation.
It therefore removes finite-sample fluctuations from the likelihood ratio used for PPO clipping while retaining latent uncertainty through $\mathbf{v}_{out}$.
As shown in Appendix~\ref{sec:gradient_variance}, policy-gradient noise comprises the intrinsic PPO variance from trajectory sampling and additional variance from finite latent sampling. MM eliminates the latter, leaving
\begin{equation}
    \text{Noise}_{total} \approx \operatorname{Var}(Y) \propto A^2 \frac{d_a}{\sigma_{act}^2 + \mathbf{v}_{out}}.
\end{equation}
This reduced-noise signal is consistent with the stable, rapidly convergent optimization observed in \cref{subsec:convergence_analysis}.
Determinism does not remove all approximation error: by discarding cross-unit covariance, MM can underestimate $\mathbf{v}_{out}$, yielding an overly narrow policy.

\subsection{Latent-Sampling Estimator}
\label{subsec:mc_estimator}

The MC estimator evaluates the actor $p_\psi$ at $N$ independent samples $z^{(i)}\sim q_\phi(\cdot\mid o)$ and averages their likelihoods:
\begin{equation}
    \begin{aligned}
    \widehat{\pi}^{\mathrm{MC}}_\theta(a\mid o)
    &= \frac{1}{N}\sum_{i=1}^{N}
    \mathcal{N}(a\mid\boldsymbol{\mu}^{(i)},\sigma_{act}^2\mathbf{I}).
    \end{aligned}
    \label{eq:mc_policy}
\end{equation}
The samples are processed in parallel by expanding the environment batch by a factor of $N$.
As $N$ grows, \cref{eq:mc_policy} consistently approaches the marginalized policy without imposing MM's diagonal-covariance propagation approximation.
It captures latent-induced correlations and trains the actor across the support of $q_\phi$, helping when observation and dynamics shifts perturb the latent representation during sim-to-real transfer~\cite{uncertainty_esti}.

Accuracy comes with a direct computational cost: actor evaluation and activation memory both scale as $\mathcal{O}(N)$.
A small $N$ leaves substantial ratio variance, while $N\geq 50$ is reliable but expensive throughout training.

\subsection{Complementary Hybrid Schedule}
\label{subsec:hybrid_schedule}

Because MM and MC share the same encoder and actor, switching between them simply replaces the marginal-policy estimator while retaining all learned parameters.
\Cref{alg:p3} summarizes this shared training interface.

\begin{algorithm}[tb]
  \caption{$P^3$ with Switchable Probability Estimators}
  \label{alg:p3}
  \begin{algorithmic}[1]
    \STATE {\bfseries Input:} Estimator schedule $\{e_k\}$, sample count $N$
    \STATE {\bfseries Initialize:} Actor $p_\psi$, VAE $q_\phi$, critic $V_\omega$
    \FOR{each PPO update $k$}
      \STATE Select $e_k$ and collect PPO data $\mathcal{D}$ with $\widehat{\pi}^{e_k}_{\theta_{old}}$
      \STATE Obtain $\boldsymbol{\mu}_z,\boldsymbol{\sigma}_z^2 \leftarrow q_\phi(o)$
      \IF{$e_k=\mathrm{MM}$}
        \STATE $(\boldsymbol{\mu}_{out},\mathbf{v}_{out}) \leftarrow
        \operatorname{MM}(p_\psi;[o,\boldsymbol{\mu}_z],[\mathbf{0},\boldsymbol{\sigma}_z^2])$
        \STATE $\widehat{\pi}^{e_k}_\theta \leftarrow
        \mathcal{N}(\boldsymbol{\mu}_{out},\operatorname{diag}(\mathbf{v}_{out})+\sigma_{act}^2\mathbf{I})$
      \ELSE
        \STATE Draw $z^{(i)}\sim q_\phi(\cdot\mid o)$ and compute
        $\boldsymbol{\mu}^{(i)}\leftarrow\operatorname{Actor}(o,z^{(i)};\psi)$ for $i=1,\ldots,N$
        \STATE $\widehat{\pi}^{e_k}_\theta(a\mid o) \leftarrow
        \frac{1}{N}\sum_{i=1}^{N}\mathcal{N}(a\mid\boldsymbol{\mu}^{(i)},\sigma_{act}^2\mathbf{I})$
      \ENDIF
      \STATE Compute \cref{eq:total_loss} using $\widehat{\pi}^{e_k}_\theta$ and $\widehat{\pi}^{e_k}_{\theta_{old}}$
      \STATE Update $\psi,\phi,\omega$
    \ENDFOR
  \end{algorithmic}
\end{algorithm}

Our default $P^3$ schedule first trains the policy with MM, whose deterministic propagation eliminates latent-sampling noise and the resulting outliers in the PPO likelihood ratio, providing a stable signal for rapid initial convergence.
After the MM policy reaches its training plateau, we switch the estimator to MC for a short \emph{Latent Sample Fine-Tuning} (LSFT) phase.
By propagating samples from $q_\phi$ through the nonlinear actor, MC more faithfully represents how the full latent distribution is transformed into the action distribution, including distributional structure that diagonal MM may omit.
LSFT builds on the well-trained MM policy and further improves its performance and robustness to latent uncertainty.

\section{Experiments}
\label{sec:experiments}
We design experiments to address the following questions:

\textbf{Q1:} To what extent can $P^3$ improve data efficiency?

\textbf{Q2:} What approximation error does MM add to the actor's output distribution, and does MC-based LSFT mitigate it?

\textbf{Q3:} Does $P^3$ improve convergence and asymptotic performance over MC estimators and other baselines?

\textbf{Q4:} Does $P^3$ improve sim-to-sim and sim-to-real transfer?

\subsection{Experimental Setup}
We evaluate $P^3$ on a challenging locomotion task: humanoid traversal of complex terrain.
Policies are trained at scale in Isaac Sim~\cite{NVIDIA_Isaac_Sim} using RSL-RL~\cite{rslrl}, then evaluated in MuJoCo~\cite{mujoco} and on a real robot.

\paragraph{Network Structure}
\label{subsec:network_structure}
We adopt an Actor-Critic framework with a VAE estimator similar to prior works ~\cite{PIM}, visualized in~\cref{fig:main_pic}.

The actor is an MLP that maps observation $\mathbf{o}_t$ and latent variable $\mathbf{z}_t$ to joint target positions $\mathbf{a}_t$.
The VAE estimator processes two inputs: proprioceptive history $\mathbf{o}^{H}_t$ and exteroceptive observation $\mathbf{o}^{\text{extero}}_t$.
Specifically, the encoder employs an MLP to encode $\mathbf{o}^{H}_t$ (simultaneously estimating velocity $\mathbf{v}_t$) and a CNN to encode $\mathbf{o}^{\text{extero}}_t$.
The decoder utilizes an MLP to reconstruct the next state $\mathbf{o}_{t+1}$ conditioned on $\mathbf{a}_t$, while a CNN reconstructs $\mathbf{o}^{\text{extero}}_t$ from the latent space.

\paragraph{Baseline and Estimator Nomenclature}
The simplest VAE architecture (similar to~\cite{PIM}) is adopted as the primary baseline to isolate confounding factors and focus on the improvements brought by probabilistic propagation.
In addition to VAE, we select the following algorithms as baselines: SimpleActorCritic~\cite{legged_gym}, SPR~\cite{spr}, and AutoEncoder (AE), detailed in Appendix~\ref{sec:baseline_detail}.
For a fair comparison, all baselines are trained using PPO-clip and the same hyperparameters. The detailed hyperparameters are listed in Appendix~\ref{sec:algorithmic_details}.
Throughout this section, \emph{MC-only} ($N$) denotes training from initialization with the MC estimator and $N$ latent samples; the VAE baseline is its single-sample case ($N=1$).
We use \emph{$P^3$-MM} for the checkpoint learned with MM before LSFT, and \emph{$P^3$} for the complete schedule, which switches that checkpoint to MC with $N=15$ for LSFT.

\paragraph{Environment and Terrain Curriculum Design}
Our simulation environment setup is similar to \cite{PIM} and \cite{beamdojo}. 
The humanoid robot is trained for locomotion on challenging terrains, including stepping stones, stairs, and gaps (detailed in Appendix~\ref{sec:environment_details}).
We adopt a widely used curriculum-based training scheme with varying terrain difficulties introduced in~\cite{attention,legged_gym}.
Robots are first assigned a random terrain type, and those who successfully traverse it are promoted to the next difficulty level, while those who fail are demoted. 
The difficulty level of each terrain is tuned heuristically; for instance, the stepping stones size decreases and sparsity increases as the curriculum progresses.

\paragraph{Hardware Setup}
Training is conducted using NVIDIA Hopper architecture GPUs. 
For real-world experiments, the model inference runs on a laptop equipped with NVIDIA RTX 5090 GPU, which transmits control commands to the G1 robot in real-time. 

The trained policy is deployed on the 29-DoF Unitree G1 humanoid robot. Robot-centric elevation map ~\cite{elevation, elevation2} receives data from Fast LiDAR-Inertial Odometry (FAST-LIO) ~\cite{fastlio, fastlio2} for Livox mid-360 radar.

\subsection{Data Efficiency Results}
\label{subsec:data_efficiency}
PPO-Clip suppresses the policy-gradient contribution of a sample once its probability ratio enters the saturated clipping branch.
Single-sample estimation can assign different probability ratios under identical policies and observation, spuriously pushing an informative sample into this branch.
We denote \emph{data efficiency} ($D_{\mathrm{eff}}$) as the fraction of samples whose estimated probability ratios remain within the clipping interval when the current and old policies are identical,
\begin{equation}
  \label{eq:data_efficiency}
  D_{\mathrm{eff}} = \frac{1}{M}\sum_{i=1}^{M}
  \mathbb{I}\!\left[1-\epsilon \leq \hat{r}_i \leq 1+\epsilon\right]
  \times 100\%.
\end{equation}
With exact marginalized-policy likelihoods, the ratio equals one for samples in this controlled setting. Any ratio outside $[1-\epsilon,1+\epsilon]$ is an estimation artifact.
This diagnostic measures the utilization of samples within a PPO update.

\begin{table}[t]
  \centering
  \begin{small}
    \begin{tabular}{lc}
      \toprule
      Probability Estimator & $D_{\mathrm{eff}}\uparrow$ \\
      \midrule
      MC ($N=1$; VAE) & 64.6\% \\
      MC ($N=5$) & 79.3\% \\
      MC ($N=15$) & 89.8\% \\
      MC ($N=50$) & 96.5\% \\
      MM ($P^3$-MM) & \textbf{100.0\%} \\
      \bottomrule
    \end{tabular}
  \end{small}
  \caption{Data efficiency under the identical-policy diagnostic. Higher values mean fewer erroneously clipped samples.}
  \label{tab:data_efficiency}
\end{table}

As reported in \cref{tab:data_efficiency}, the single-sample VAE retains only 64.6\% of samples within the unclipped interval, meaning that 35.4\% are spuriously exposed to clipping.
Increasing the number of latent samples progressively improves the MC estimate, reaching 96.5\% data efficiency at $N=50$; by contrast, MM attains $D_{\mathrm{eff}}=100.0\%$ by construction.

This data efficiency loss can be traced to KL estimation error because the PPO ratio and policy KL divergence satisfy
$D_{KL}(\pi_{\theta_{old}}\|\pi_\theta)=\mathbb{E}_{a\sim\pi_{\theta_{old}}}[-\log r_\theta(a)]$ and share the same estimated marginalized-policy likelihoods.
If finite-sample likelihood error drives erroneous clipping, its KL estimate should approach a stable reference as $N$ grows.

\begin{figure}[t]
  \centering
  \includegraphics[width=\columnwidth]{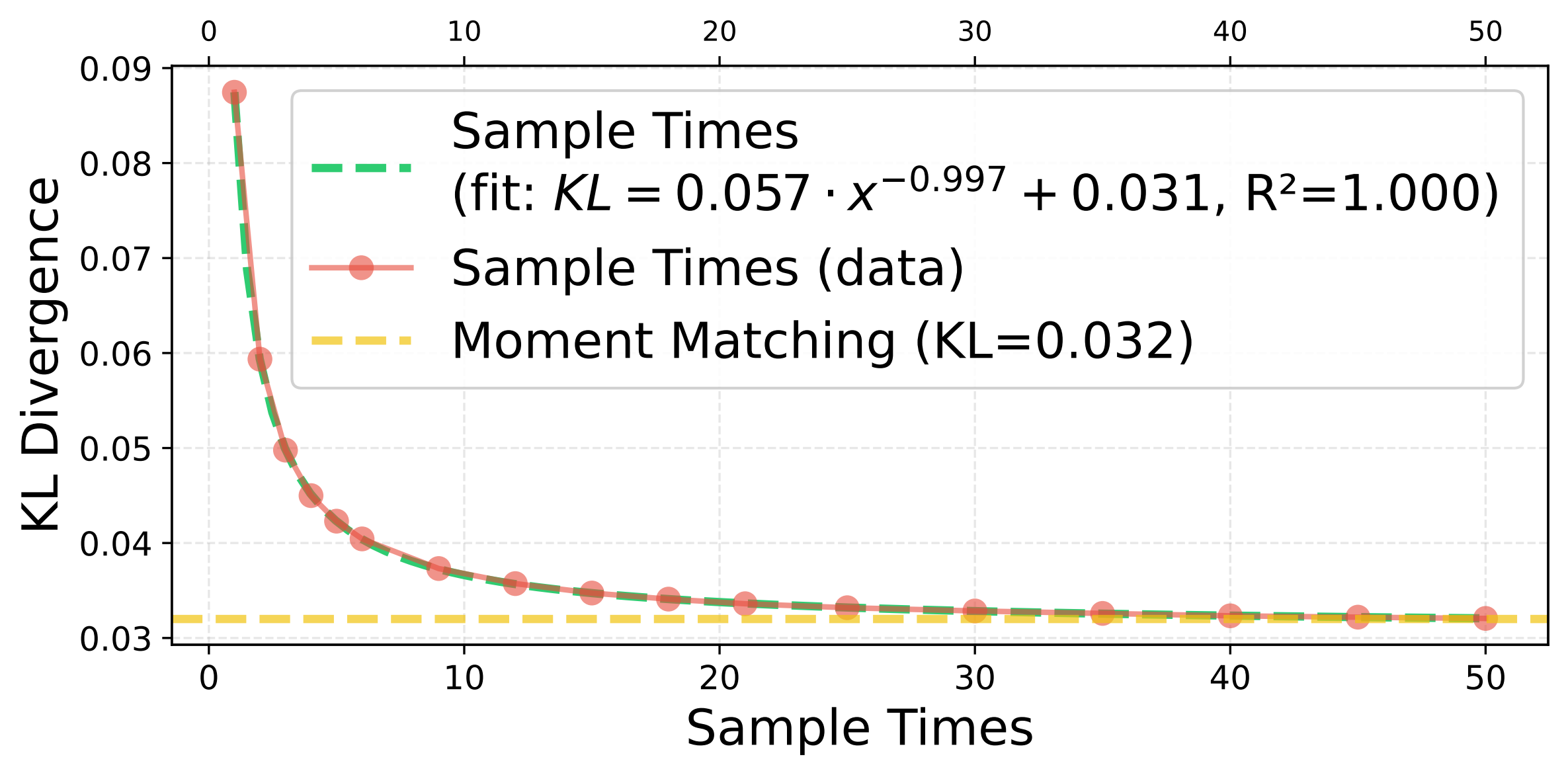}
  \caption{
    \textbf{KL-divergence estimates for a single-epoch update at epoch 1000.}
    MC approaches a stable high-sample reference as $N$ increases, while MM achieves comparable accuracy without latent sampling.
  }
  \label{kl_scan_comparison}
\end{figure}

\Cref{kl_scan_comparison} confirms this prediction: the single-sample VAE overestimates the stable high-sample KL value by approximately a factor of three, whereas the discrepancy decreases with $N$ and MM closely matches the estimate obtained with more than 30 samples.
Together with \cref{tab:data_efficiency}, this result shows that single-sample KL bias causes erroneous clipping and reduces data utilization.

\subsection{Action Distribution Results}
\label{subsec:action_analysis}
We compare the action distribution approximated by MM with an empirical reference obtained by direct latent sampling.
As illustrated in Figure~\ref{fig:double_dim_dist}, with further analysis in Appendix~\ref{subsec:action_distribution}, the actor's output resembles a multivariate Gaussian distribution with non-diagonal covariance, indicating linear correlations between dimensions.

MM propagates diagonal first- and second-order statistics and therefore does not explicitly model cross-dimensional covariance.
At the $P^3$-MM checkpoint, the top panel of \cref{fig:double_dim_dist} shows that its propagated distribution underestimates the action variance relative to the MC reference while preserving a closely aligned action mean.
This mean fidelity provides a low-variance  optimization signal for learning.

After LSFT, the bottom panel shows closer agreement between the propagated and sampled distributions, indicating that LSFT mitigates MM's variance underestimation.
\begin{figure}[t]
  \centering
  \includegraphics[width=0.8\columnwidth]{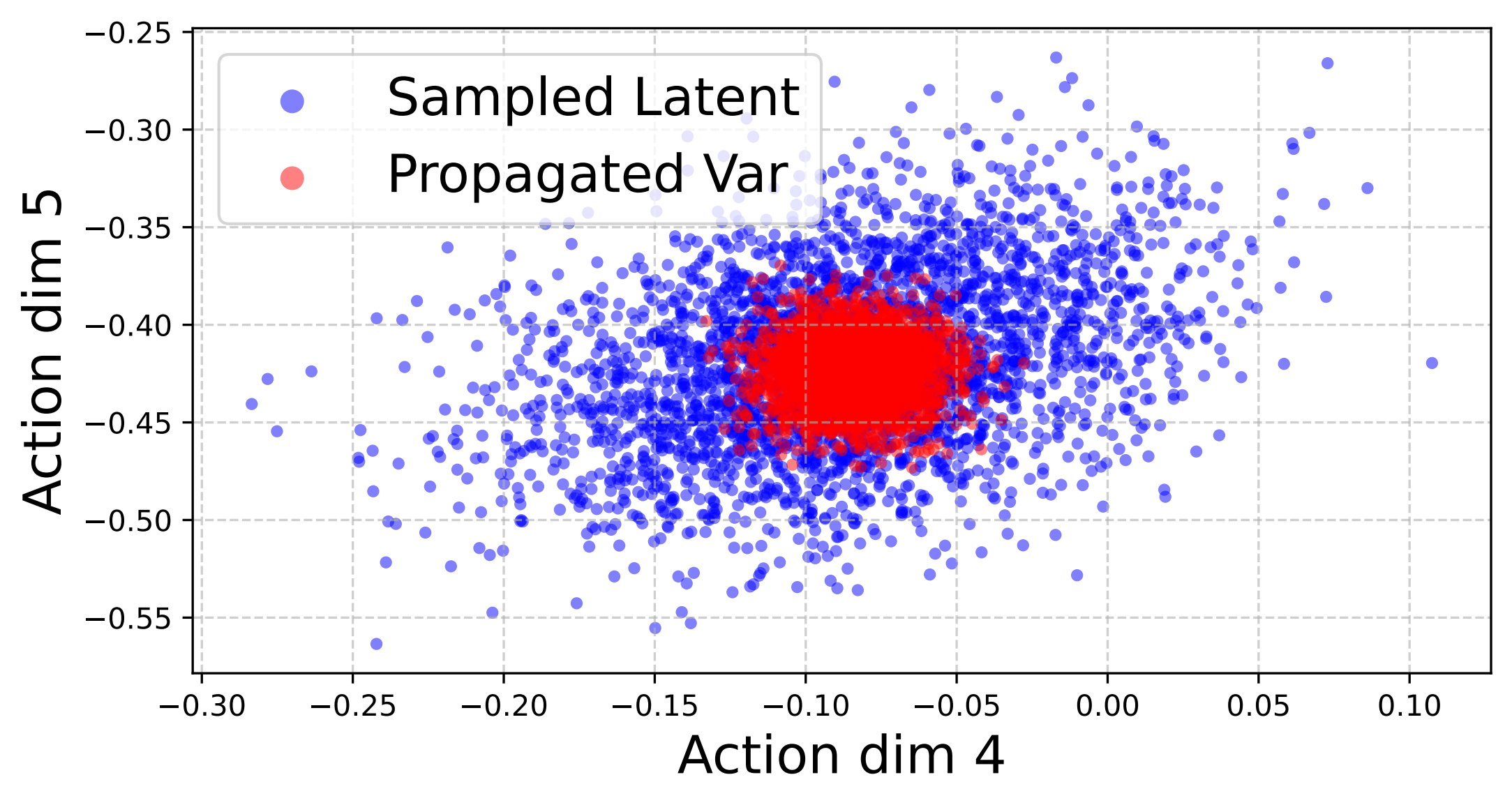}

  \includegraphics[width=0.8\columnwidth]{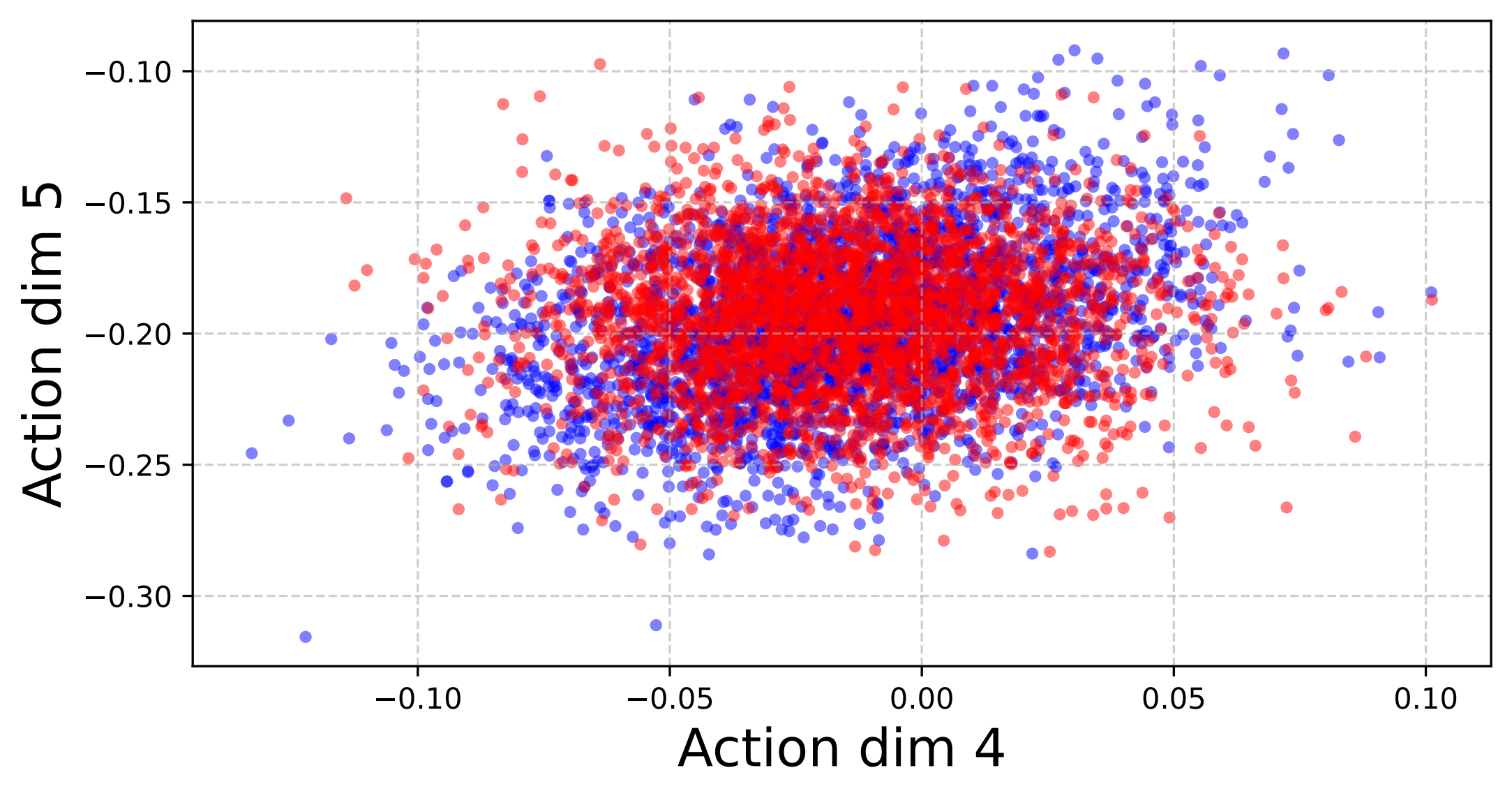}
  \caption{
    \textbf{Action-distribution approximation in $P^3$.}
    Blue points are actor outputs from direct latent sampling (the MC reference), and red points are drawn from the distribution obtained by MM propagation.
    The top panel uses the $P^3$ checkpoint before LSFT; the bottom panel uses it after LSFT.
  }
  \label{fig:double_dim_dist}
\end{figure}

\subsection{Convergence Analysis}
\label{subsec:convergence_analysis}
\paragraph{Evaluation Metrics}
We report \emph{curriculum difficulty}, the mean terrain-curriculum level reached by the training population, as the primary learning-progress metric.
A higher value indicates reliable traversal of harder terrain.

\paragraph{Comparison of Probability Estimators}
\label{subsec:prob_method}

\begin{figure}[t]
  \centering
  \includegraphics[width=\columnwidth]{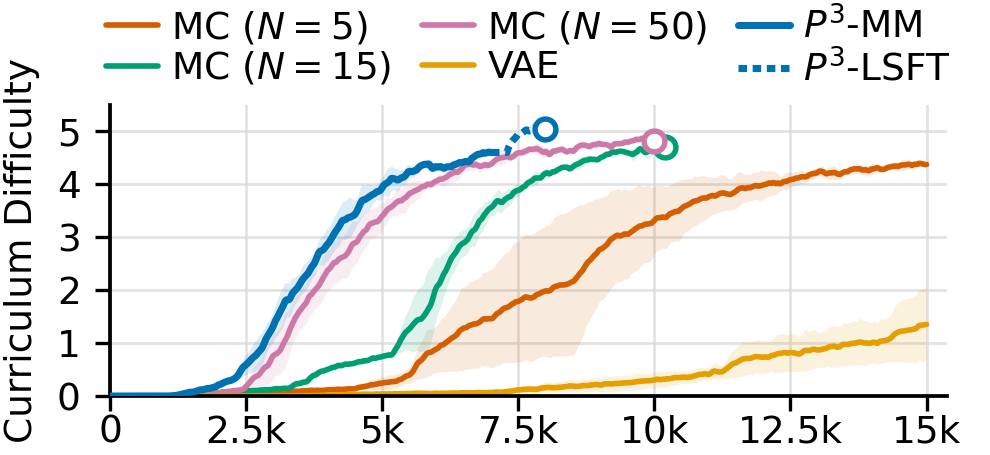}
  \caption{
    \textbf{Comparison of probability estimators.}
    Curriculum difficulty is plotted against PPO training epochs.
    Solid curves and bands denote the mean and one standard deviation; the dashed segment is a dedicated 1,000-epoch LSFT ($N=15$).
    Open circles indicate convergence.
  }
  \label{convergence_ada_pic}
\end{figure}

For all curves, a successful stopping point requires a sustained near-asymptotic plateau and a terminal curriculum difficulty of at least 4.5; methods that fail this criterion remain visible through 15,000 epochs.
Under this protocol, $P^3$ reaches a high-difficulty plateau after 7,000 epochs of MM training. The complete $P^3$ schedule then switches to MC with $N=15$ for 1,000 epochs of LSFT.

MC-only ($N=50$), the strongest sampling baseline, converges at epoch 10,000 and at a lower difficulty; $P^3$ therefore requires $20\%$ fewer training epochs. MC-only ($N=15$) converges slightly later. Neither MC-only ($N=5$) nor the single-sample VAE converges within the 15,000-epoch budget; both exhibit pronounced training instability, consistent with the high variance of their finite-sample probability estimates. Although increasing $N$ improves performance, it expands actor evaluation and memory cost proportionally to the sample count (Appendix~\ref{subsec:computational_cost}).

This comparison isolates the roles of the two estimators: MM efficiently acquires a strong policy, after which a short MC-based LSFT phase improves robustness to latent uncertainty. Their combination gives $P^3$ a more favorable optimization trajectory than MC-only training.

\paragraph{Comparison with Other Algorithms}
\label{subsec:compare_baseline}
\begin{figure}[t]
  \centering
  \includegraphics[width=\columnwidth]{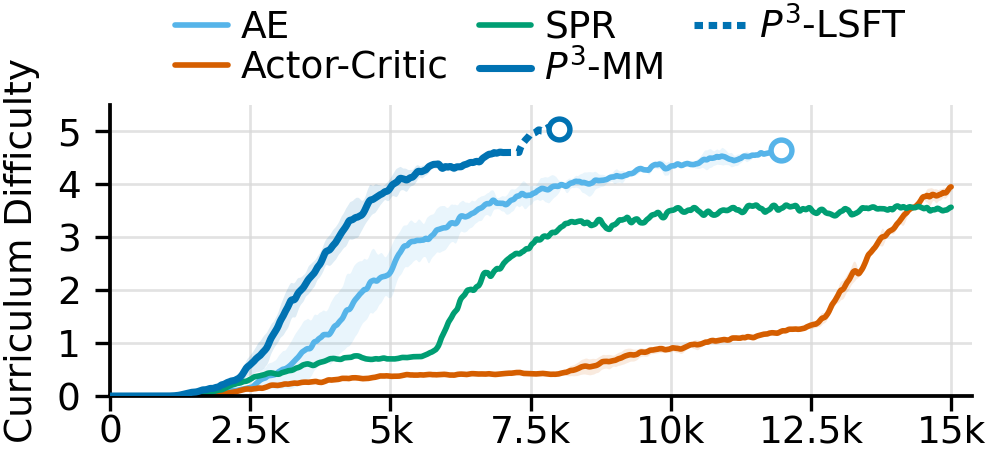}
  \caption{
    \textbf{Comparison with representation-learning baselines.}
    Curves follow the same notation and stopping protocol as \cref{convergence_ada_pic}.
    The $P^3$ trajectory comprises 7,000 epochs with MM followed by 1,000 epochs of MC-based LSFT ($N=15$).
  }
  \label{convergence_compare}
\end{figure}
\Cref{convergence_compare} compares the $P^3$ training trajectory with AE, SPR, and a direct Actor--Critic baseline.
$P^3$ converges first, at epoch 8,000, and attains the highest terminal curriculum difficulty.
AE is the only other method to converge, reaching a lower terminal level (approximately 4.65) at epoch 12,000; thus, $P^3$ requires $33\%$ fewer training epochs.
SPR and Actor--Critic do not satisfy the criterion within 15,000 epochs.

Together with the controlled clipping diagnostic in \cref{tab:data_efficiency}, these results link improved probability estimation to an observable optimization benefit: $P^3$ both reaches difficult terrain earlier and finishes at a stronger policy checkpoint than the competing estimators and architectures.

\subsection{Performance Analysis}
\label{subsec:performance_analysis}

To evaluate the performance and robustness of our policy, we reconstructed the stepping stones, stairs, and gap terrains from IsaacLab within the MuJoCo physics engine.
MuJoCo employs a convex optimization formulation for contact dynamics, yielding high-precision physical interactions that differ naturally from the dynamics in IsaacLab~\cite{mujoco}.
This discrepancy serves as a rigorous test for the policy's transfer capabilities.

\begin{table}[!b]
  \centering
  \begin{small}
    \begin{sc}
      \begin{tabular}{lcc}
        \toprule
        Method & Total Reward$\uparrow$ & Lifetime$\uparrow$ \\
        \midrule
        VAE & 16.2 & 15.4 \\
        SimpleActorCritic & 13.7 & 18.3 \\
        AE & 17.9 & 18.0 \\
        SPR & 10.7 & 14.4 \\
        MC-only ($N=50$) & 18.2 & 19.4 \\
        $P^3$-MM & 18.9 & 19.7 \\
        $P^3$ & \textbf{20.1} & \textbf{20.0} \\
        \bottomrule
      \end{tabular}
    \end{sc}
  \end{small}
  \caption{Performance comparison in MuJoCo.}
  \label{tab:performance}
\end{table}

Table~\ref{tab:performance} presents a performance comparison of the different algorithms in MuJoCo.
All models use either the most recent checkpoint after convergence or the 15,000-epoch result if not converged.
The $P^3$-MM policy is evaluated at the 7,000-epoch checkpoint; the complete $P^3$ policy continues from that checkpoint for 1,000 epochs of LSFT with $N=15$ and terminates at epoch 8,000.

The deterministic AE baseline outperforms the stochastic VAE in both total reward (17.9 vs.\ 16.2) and lifetime (18.0 vs.\ 15.4), whereas SPR exhibits the weakest transfer performance. 
More importantly, $P^3$-MM already surpasses MC-only ($N=50$) on both metrics at its 7,000-epoch checkpoint.
LSFT further increases the total reward from 18.9 to 20.1 and the lifetime from 19.7 to 20.0. 
Consequently, the complete $P^3$ achieves the best overall performance in MuJoCo across both metrics.

\begin{figure}[!t]
  \centering
  \includegraphics[width=\columnwidth]{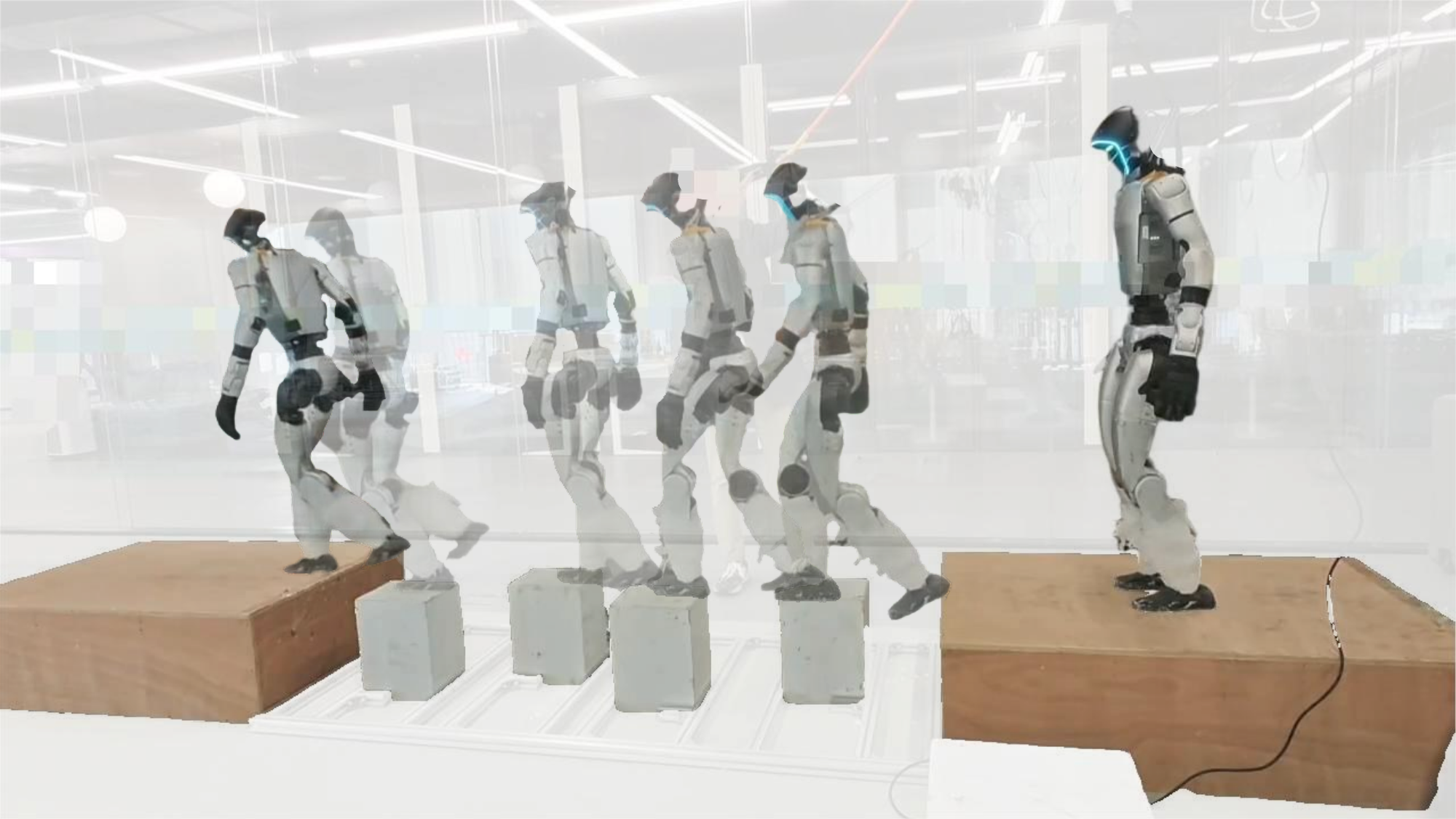}
  \includegraphics[width=\columnwidth]{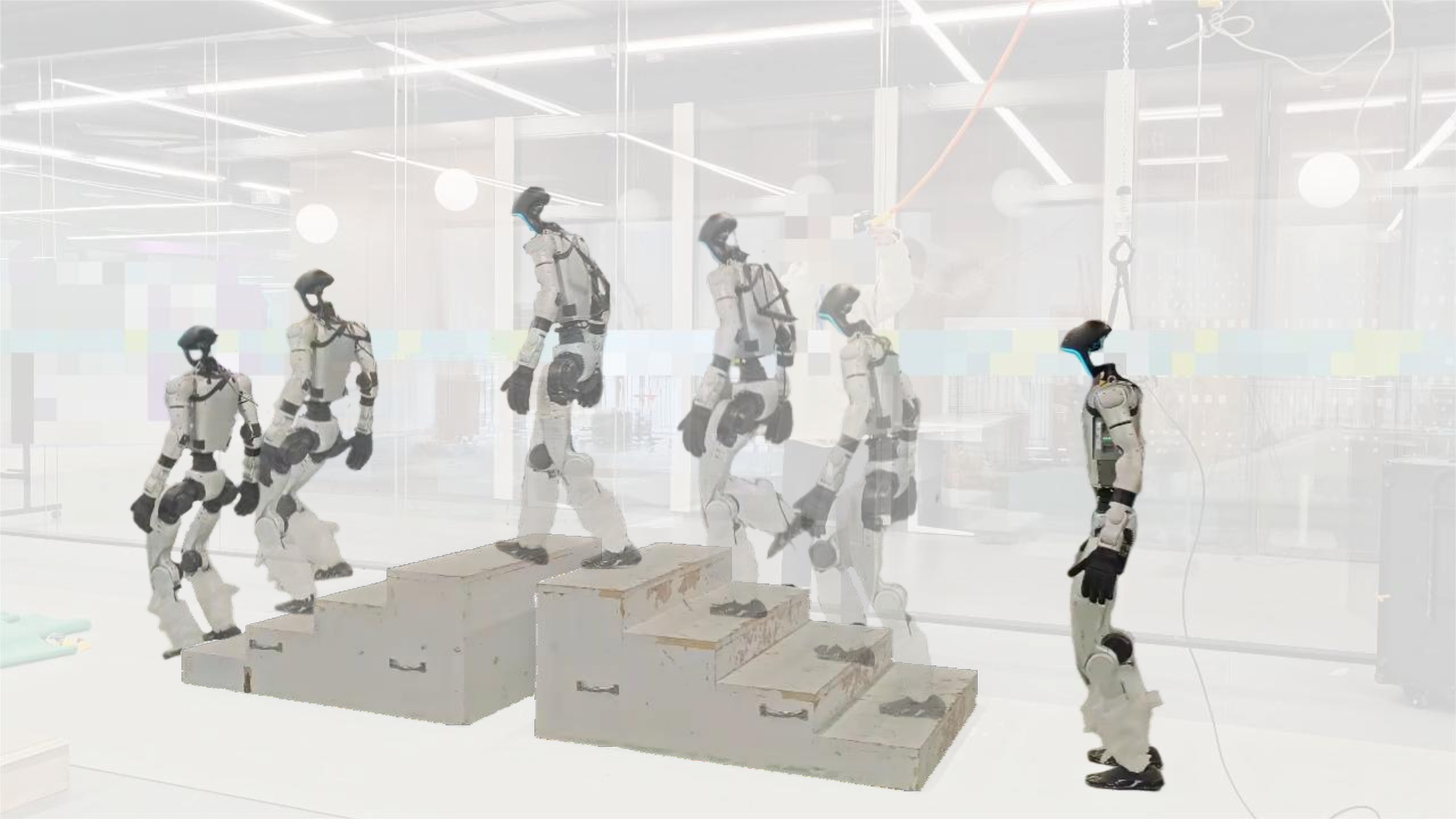}
  \caption{
    \textbf{Real-world evaluation.}
    We validated the effectiveness of $P^3$ on the G1 robot in the real world. The robot successfully navigated through challenging terrains such as stepping stones and stairs.
  }
  \label{fig:exp_pic}
\end{figure}

\begin{table}[t]
  \centering
  \begin{small}
    \begin{sc}
      \begin{tabular}{lccc}
        \toprule
        Method & Stepping & Stairs & Gaps \\
        \midrule
        VAE & 6 & 7 & 7 \\
        SimpleActorCritic & 0 & 2 & 3 \\
        AE & 4 & 7 & 7 \\
        SPR & 2 & 2 & 4 \\
        MC-only ($N=50$) & \textbf{8} & 7 & 9 \\
        $P^3$-MM & 7 & 7 & 9 \\
        $P^3$ & \textbf{8} & \textbf{9} & \textbf{10} \\
        \bottomrule
      \end{tabular}
    \end{sc}
  \end{small}
  \caption{Success counts in real-world deployment (10 trials).}
  \label{tab:real_performance}
\end{table}
Real-world experiments, as summarized in~\cref{tab:real_performance}, demonstrate the effectiveness of our method in deployment.
Despite elevation-map drift and kinematic discrepancies, $P^3$ attains the highest overall success counts in \cref{fig:exp_pic}.

\FloatBarrier
\section{Conclusion}
\label{sec:conclusion}

This work targets a specific yet consequential bottleneck in a widely adopted robotics framework: optimizing a stochastic VAE state estimator together with PPO.
We propose Probabilistic Policy Propagation ($P^3$), enabling this architecture to retain VAE's advantages while being optimized as a coherent policy distribution.
By removing a long-standing optimization barrier without abandoning this proven robotics paradigm, $P^3$ turns VAE-based policy learning into a more reliable and scalable foundation, opening the door to its broader adoption across robots and sensing modalities.

\bibliography{src/references}

\clearpage
\appendix
\twocolumn[
  \centering
  {\LARGE\bfseries Appendix\par}
  \vspace{1.5em}
]
\section{Additional Theoretical Analysis}
\label{sec:additional_theory}

We follow the notation of the main paper: $q_\phi(z\mid o)$ is the VAE
encoder, $p_\psi(a\mid z)$ is the latent-conditioned actor,
$\theta=(\phi,\psi)$ denotes their joint parameters, and $N$ is the number of
latent samples used by a Monte Carlo (MC) estimator. The effective policy is
the marginalized distribution
$\pi_\theta(a\mid o)=\int p_\psi(a\mid z)q_\phi(z\mid o)\,dz$.

\subsection{KL Divergence Decomposition}
\label{sec:kl_de}
We analyze the decomposition of the policy KL divergence. 
Substituting the hierarchical policy definition $\pi_\theta(a|o) = \mathbb{E}_{z \sim q_\phi(z|o)}[p_\psi(a|z)]$ into the standard KL divergence, we derive a tractable upper bound.
We apply Jensen's inequality utilizing the convexity of the function $f(x) = -\log(x)$, which implies $f(\mathbb{E}[x]) \le \mathbb{E}[f(x)]$. 
This bound becomes an equality if and only if the random variable inside the logarithm is constant almost everywhere. 
For compactness, let $P_{old}(a,z|o)=p_{\psi_{old}}(a|z)q_{\phi_{old}}(z|o)$ and $P(a,z|o)=p_\psi(a|z)q_\phi(z|o)$.
The derivation proceeds as follows:
\begin{equation}
\begin{aligned}
  &D_{KL}(\pi_{\theta_{old}}(\cdot|o)\parallel\pi_\theta(\cdot|o)) \\
  &\le D_{KL}(P_{old}(a,z|o)\parallel P(a,z|o)) \\
  &= \iint P_{old}(a,z|o)
     \log\frac{p_{\psi_{old}}(a|z)}{p_\psi(a|z)}\,dz\,da \\
  &\quad + \iint P_{old}(a,z|o)
     \log\frac{q_{\phi_{old}}(z|o)}{q_\phi(z|o)}\,dz\,da \\
  &= \mathbb{E}_{z\sim q_{\phi_{old}}(\cdot|o)}
     \!\left[D_{KL}\!\left(p_{\psi_{old}}(\cdot|z)\parallel p_\psi(\cdot|z)\right)\right] \\
  &\quad + D_{KL}\!\left(q_{\phi_{old}}(\cdot|o)\parallel q_\phi(\cdot|o)\right).
\end{aligned}
\end{equation}
This is the detailed form of the upper bound used in the main paper. It
separates the shift of the latent estimator from the change of the actor over
the full old latent distribution, rather than at only one sampled latent.

\subsection{Variance of the Probability Ratio}
\label{sec:variance_ratio}

We quantitatively analyze the variance introduced into the surrogate loss by the stochastic sampling process. 
We define the actor policy with PPO's exploration variance as $p_\psi(a|z) = \mathcal{N}(\mu_\psi(z), \sigma_{act}^2 I_{d_a})$.
Consider an $N$-sample Monte Carlo (MC) estimator for the probability ratio:
\begin{equation}
    \hat{r}_N = \frac{\frac{1}{N}\sum_{i=1}^N p_\psi(a|z_i)}{\frac{1}{N}\sum_{j=1}^N p_{\psi_{old}}(a|z'_j)}.
\end{equation}
Writing the numerator and denominator as the independent sample-mean estimates $\bar{X}$ and $\bar{Y}$, respectively, gives $\hat{r}_N=\bar{X}/\bar{Y}$. The Delta Method provides the following first-order variance approximation:
\begin{equation}
    \text{Var}\left(\frac{\bar{X}}{\bar{Y}}\right) \approx \left(\frac{\mathbb{E}[\bar{X}]}{\mathbb{E}[\bar{Y}]}\right)^2 \left( \frac{\text{Var}(\bar{X})}{\mathbb{E}[\bar{X}]^2} + \frac{\text{Var}(\bar{Y})}{\mathbb{E}[\bar{Y}]^2} \right).
\end{equation}
Noting that $\text{Var}(\bar{X}) = \frac{1}{N}\text{Var}(X)$ and $\mathbb{E}[\hat{r}_N] \approx r_{\mathrm{true}}$, we obtain:
\begin{equation}
    \text{Var}(\hat{r}_N) \approx \frac{r_{\mathrm{true}}^2}{N} \left( \underbrace{\frac{\text{Var}(p_\psi)}{\mathbb{E}[p_\psi]^2}}_{CV^2_{\psi}} + \underbrace{\frac{\text{Var}(p_{\psi_{old}})}{\mathbb{E}[p_{\psi_{old}}]^2}}_{CV^2_{\psi_{old}}} \right),
\end{equation}
where $CV$ denotes the coefficient of variation of the policy probability density induced by latent sampling.

\subsection{Variance Formula Expansion}
\label{sec:variance_expansion}

We derive the analytical form of $CV^2_\psi$. Let $z \in \mathbb{R}^{d_z}$ and $a \in \mathbb{R}^{d_a}$. 
The VAE latent space follows a diagonal Gaussian distribution with covariance $\Sigma_{vae} = \sigma_{vae}^2 I_{d_z}$. 
As empirically reported in the main paper's action-distribution results, the actor's output under latent sampling exhibits characteristics of a multivariate Gaussian distribution.
This observation implies that the actor network behaves approximately linearly within the variance range of the latent space. 
We further validate this linearity assumption through a dedicated regression test (\cref{subsec:action_distribution}), which yields an average $R^2$ of 0.9706 across all action dimensions. 
These results justify the use of a first-order Taylor expansion of $p_\psi(a|z)$ with respect to $z$ around the mean $\bar{z}$:
\begin{equation}
    p_\psi(a|z) \approx p_\psi(a|\bar{z}) + \nabla_z p_\psi(a|\bar{z})^\top (z - \bar{z}).
\end{equation}
The variance of $p_\psi(a|z)$ induced by $z$ is approximately:
\begin{equation}
    \text{Var}_z(p_\psi) \approx \nabla_z p_\psi^\top \Sigma_{vae} \nabla_z p_\psi = \sigma_{vae}^2 \|\nabla_z p_\psi\|^2.
\end{equation}
Using the identity $\nabla_z p = p \nabla_z \log p$, we have $CV^2_\psi = \frac{\text{Var}(p)}{\mathbb{E}[p]^2} \approx \sigma_{vae}^2 \|\nabla_z \log p_\psi\|^2$.
For the Gaussian policy, $\nabla_z \log p_\psi(a|z) = \nabla_z \left( -\frac{1}{2\sigma_{act}^2} \|a - \mu_\psi(z)\|^2 \right) = \frac{1}{\sigma_{act}^2} \mathbf{J}_\psi^\top (a - \mu_\psi(z))$, where $\mathbf{J}_\psi \in \mathbb{R}^{d_a \times d_z}$ is the Jacobian of the actor mean $\mu_\psi(z)$.
Substituting this back, we get:
\begin{equation}
    CV^2_\psi \approx \frac{\sigma_{vae}^2}{\sigma_{act}^4} (a - \mu_\psi(\bar{z}))^\top \mathbf{J}_\psi \mathbf{J}_\psi^\top (a - \mu_\psi(\bar{z})).
\end{equation}
Since $\hat{r}_N$ involves samples of actions drawn from the policy, we take the expectation of this term over $a \sim p_\psi(a|\bar{z})$. Using $\mathbb{E}[(a-\mu)(a-\mu)^\top] = \sigma_{act}^2 I_{d_a}$, we find:

\begin{equation}
    \mathbb{E}_a [CV^2_\psi] \approx \frac{\sigma_{vae}^2}{\sigma_{act}^4} \text{Tr}\left( \mathbf{J}_\psi \mathbf{J}_\psi^\top \sigma_{act}^2 I_{d_a} \right) = \frac{\sigma_{vae}^2}{\sigma_{act}^2} \|\mathbf{J}_\psi\|_F^2.
\end{equation}
Substituting this back into the expression for $\text{Var}(\hat{r}_N)$, we have the scaling relation
\begin{equation}
    \text{Var}(\hat{r}_N) \propto \frac{r_{\mathrm{true}}^2}{N} \cdot \left( \frac{\sigma_{vae}^2 \|\mathbf{J}_\psi\|_F^2}{\sigma_{act}^2} \right).
\end{equation}
Near a PPO update, the current and old policies have comparable latent
variances and local actor Jacobians. Consequently, both coefficient-of-
variation terms in \cref{sec:variance_ratio} have the same scaling; the last
display absorbs their constant factor and retains the dependence on $N$,
$\sigma_{vae}$, $\sigma_{act}$, and $\|\mathbf{J}_\psi\|_F$ emphasized in the
main paper.

\subsection{Variance of the Gradient}
\label{sec:gradient_variance}

We analyze how the variance in the probability ratio propagates to the policy gradient. 
Note that $\nabla_\theta r_\theta = \nabla_\theta \left( \frac{\pi_\theta}{\pi_{\theta_{old}}} \right) = \frac{\pi_\theta}{\pi_{\theta_{old}}} \nabla_\theta \log \pi_\theta = r_\theta \nabla_\theta \log \pi_\theta$.
Thus, the gradient can be written as the product of the advantage, the noisy ratio, and the score function:
\begin{equation}
\label{eq:grad_prod}
    \hat{g} \approx A \cdot \hat{r}_N \cdot \nabla_\theta \log \pi_\theta.
\end{equation}
Conditioned on a fixed state-action pair $(s, a)$ from the PPO rollout buffer, the term $Y = A \cdot \nabla_\theta \log \pi_\theta$ is deterministic, while the variance of $\hat{r}_N$ arises solely from latent sampling.
Conversely, the variance of $Y$ across the batch stems from trajectory sampling, which is independent of the latent sampling process. 
Therefore, we treat $X = \hat{r}_N$ and $Y$ as conditionally independent and apply the product variance formula: $\text{Var}(XY) \approx \mathbb{E}[X]^2 \text{Var}(Y) + \mathbb{E}[Y]^2 \text{Var}(X) + \text{Var}(X)\text{Var}(Y)$.
Since $\mathbb{E}[\hat{r}_N] \approx 1$, the gradient variance is
\begin{equation}
\begin{aligned}
  \operatorname{Var}(\hat{g})
  &\approx \operatorname{Var}(Y)
    + \mathbb{E}[Y]^2\operatorname{Var}(\hat{r}_N) \\
  &\quad + \operatorname{Var}(\hat{r}_N)\operatorname{Var}(Y) \\
  &= \operatorname{Var}(Y)
    + \operatorname{Var}(\hat{r}_N)\mathbb{E}[\|Y\|^2].
\end{aligned}
\end{equation}

Here, $\text{Var}(Y)$ represents the intrinsic sampling variance of the policy gradient (standard PPO variance), while the second term represents the additional noise introduced by VAE latent sampling.
Under the approximately Gaussian distribution of the actor mean $\mu_\psi(z)$ established above, the marginalized policy $\pi_\theta(a)=\int q_\phi(z\mid o)p_\psi(a\mid z)\,dz$ is approximated by convolving that action-mean distribution with the actor's Gaussian exploration noise.
The second moment $\mathbb{E}[\|Y\|^2]$ is proportional to $A^2\mathbb{E}[\|\nabla\log p_\psi\|^2]$. For a Gaussian policy, the trace of the Fisher information matrix is $\mathbb{E}[\|\nabla_\mu\log p_\psi\|^2]=d_a/\sigma_{act}^2$. Combining this identity with $\text{Var}(\hat{r}_N)$ from \cref{sec:variance_expansion}, the latent-induced noise scales as
\begin{equation}
\begin{aligned}
  \text{Noise}_{latent}
  &\propto \frac{A^2}{N}
    \underbrace{\frac{d_a}{\sigma_{act}^2}}_{\text{Score variance}} \\
  &\quad \times
    \underbrace{\frac{\sigma_{vae}^2\|\mathbf{J}_\psi\|_F^2}
                      {\sigma_{act}^2}}_{\text{Ratio variance}} \\
  &= \frac{A^2}{N}
    \frac{d_a\sigma_{vae}^2\|\mathbf{J}_\psi\|_F^2}{\sigma_{act}^4}.
\end{aligned}
\end{equation}

\subsection{Derivation of Probabilistic Activation Moments}
\label{sec:elu_derivation}

Moment matching represents each pre-activation by a marginal Gaussian and
applies the following scalar formulas elementwise. Let
$x\sim\mathcal{N}(\mu,\sigma^2)$, and let $\phi(\cdot)$ and $\Phi(\cdot)$
denote the standard normal PDF and CDF, respectively. We define the
standardized threshold $\beta=\mu/\sigma$. We first derive the moments of the
ELU activation used in our experiments and then give the corresponding
closed-form updates for ReLU and Leaky ReLU.

\subsubsection{ELU}
The ELU activation is defined as
\begin{equation}
    f(x) = \begin{cases} x & x > 0 \\ \alpha (e^x - 1) & x \le 0 \end{cases}
\end{equation}

\paragraph{Preliminaries}
We utilize the following identity for the truncated expectation of an exponential function under a Gaussian distribution. For any constant $k$:
\begin{equation}
\label{eq:gauss_exp_identity}
    \int_{-\infty}^{0} e^{kx} \mathcal{N}(x|\mu, \sigma^2) \, dx = e^{k\mu + \frac{1}{2}k^2\sigma^2} \Phi\left(-\beta - k\sigma\right).
\end{equation}
\textit{Proof:} Standardize $x = \sigma t + \mu$. The integral becomes $e^{k\mu} \int_{-\infty}^{-\beta} e^{k\sigma t} \phi(t) \, dt$. Completing the square in the exponent of the integrand yields $e^{k\sigma t - t^2/2} = e^{\frac{1}{2}k^2\sigma^2} e^{-(t-k\sigma)^2/2}$. The remaining integral is over $\mathcal{N}(k\sigma, 1)$ up to $-\beta$, which evaluates to $\Phi(-\beta - k\sigma)$.

\paragraph{First Moment (Expectation).}
The expectation $\mathbb{E}[y]$ decomposes into linear and exponential regions:
\begin{equation}
\begin{aligned}
  \mathbb{E}[y]
  &= \int_{0}^{\infty} x\mathcal{N}(x|\mu,\sigma^2)\,dx \\
  &\quad + \alpha\int_{-\infty}^{0}(e^x-1)
     \mathcal{N}(x|\mu,\sigma^2)\,dx.
\end{aligned}
\end{equation}
1. \textbf{Linear Part ($x > 0$):} Using standard rectified Gaussian results:
\begin{equation}
    \int_{0}^{\infty} x \mathcal{N}(x|\mu, \sigma^2) \, dx = \mu \Phi(\beta) + \sigma \phi(\beta).
\end{equation}
2. \textbf{Exponential Part ($x \le 0$):} Using Identity \eqref{eq:gauss_exp_identity} with $k=1$ and $k=0$:
\begin{equation}
    \int_{-\infty}^{0} (e^x - 1) \mathcal{N}(x|\mu, \sigma^2) \, dx = e^{\mu + \frac{\sigma^2}{2}} \Phi(-\beta - \sigma) - \Phi(-\beta).
\end{equation}
Combining these yields the mean:
\begin{equation}
    \mathbb{E}[y] = \mu \Phi(\beta) + \sigma \phi(\beta) + \alpha \left[ e^{\mu + \frac{\sigma^2}{2}} \Phi(-\beta - \sigma) - \Phi(-\beta) \right].
\end{equation}

\paragraph{Second Moment.}
The second raw moment $\mathbb{E}[y^2]$ is similarly decomposed:
\begin{equation}
\begin{aligned}
  \mathbb{E}[y^2]
  &= \int_{0}^{\infty}x^2\mathcal{N}(x|\mu,\sigma^2)\,dx \\
  &\quad + \alpha^2\int_{-\infty}^{0}(e^x-1)^2
     \mathcal{N}(x|\mu,\sigma^2)\,dx.
\end{aligned}
\end{equation}
1. \textbf{Linear Part:} For the second moment of a rectified Gaussian:
\begin{equation}
    \int_{0}^{\infty} x^2 \mathcal{N}(x|\mu, \sigma^2) \, dx = (\mu^2 + \sigma^2)\Phi(\beta) + \mu\sigma\phi(\beta).
\end{equation}
2. \textbf{Exponential Part:} Expanding $(e^x - 1)^2 = e^{2x} - 2e^x + 1$ and applying Identity \eqref{eq:gauss_exp_identity} for $k=2, 1, 0$:
\begin{equation}
\begin{aligned}
  &\int_{-\infty}^{0}(e^{2x}-2e^x+1)
    \mathcal{N}(x|\mu,\sigma^2)\,dx \\
  &\quad = e^{2\mu+2\sigma^2}\Phi(-\beta-2\sigma) \\
  &\qquad -2e^{\mu+\frac{\sigma^2}{2}}\Phi(-\beta-\sigma)
    +\Phi(-\beta).
\end{aligned}
\end{equation}
Combining these, we get:
\begin{equation}
\begin{aligned}
  \mathbb{E}[y^2]
  &= (\mu^2+\sigma^2)\Phi(\beta)+\mu\sigma\phi(\beta) \\
  &\quad +\alpha^2\Big[
     e^{2\mu+2\sigma^2}\Phi(-\beta-2\sigma) \\
  &\qquad -2e^{\mu+\frac{\sigma^2}{2}}\Phi(-\beta-\sigma)
     +\Phi(-\beta)\Big].
\end{aligned}
\end{equation}
The propagated variance used by the moment-matching (MM) estimator is then
\begin{equation}
    \operatorname{Var}[y]=\mathbb{E}[y^2]-\mathbb{E}[y]^2.
\end{equation}
Thus, the expressions above provide the complete deterministic ELU update for
the first two moments used by $P^3$-MM.

\subsubsection{ReLU}
The MM estimator is not specific to ELU: it only requires an
activation-specific map from input moments to output moments. To express the
updates in the notation of the main paper, let
$\mu_{in}=\mu$, $v_{in}=\sigma^2$, and
$\beta=\mu_{in}/\sqrt{v_{in}}$ for $v_{in}>0$.

For the ReLU activation $y=\max(0,x)$, the positive-half Gaussian moments give
\begin{equation}
\begin{aligned}
  \mu_{out}
  &= \mu_{in}\Phi(\beta)
     +\sqrt{v_{in}}\,\phi(\beta), \\
  v_{out}
  &= (\mu_{in}^2+v_{in})\Phi(\beta)
     +\mu_{in}\sqrt{v_{in}}\,\phi(\beta)
     -\mu_{out}^2.
\end{aligned}
\label{eq:relu_moments}
\end{equation}

\subsubsection{Leaky ReLU}
For Leaky ReLU,
$y=x$ when $x>0$ and $y=\lambda x$ when $x\leq 0$, where $\lambda$ is the
negative-slope coefficient. Its propagated moments are
\begin{equation}
\begin{aligned}
  \mu_{out}
  &= \mu_{in}\big[\Phi(\beta)+\lambda\Phi(-\beta)\big] \\
  &\quad +(1-\lambda)\sqrt{v_{in}}\,\phi(\beta), \\
  v_{out}
  &= (\mu_{in}^2+v_{in})
     \big[\Phi(\beta)+\lambda^2\Phi(-\beta)\big] \\
  &\quad +(1-\lambda^2)\mu_{in}\sqrt{v_{in}}\,\phi(\beta)
     -\mu_{out}^2.
\end{aligned}
\label{eq:leaky_relu_moments}
\end{equation}
Setting $\lambda=0$ recovers the ReLU formulas in
\cref{eq:relu_moments}, while $\lambda=1$ recovers the identity map. Hence,
the layerwise MM construction extends directly to these common
piecewise-linear activations. When $v_{in}=0$, each update is understood in
the deterministic limit, $\mu_{out}=f(\mu_{in})$ and $v_{out}=0$. We use ELU
in the reported experiments to keep the actor architecture fixed across all
comparisons, rather than because $P^3$ requires a particular activation
function.

\section{Additional Experimental Setup}
\label{sec:experiment_setup}

\subsection{Network Architecture Details}
\label{sec:network_architectures}

We use the same Actor--Critic framework with a VAE state estimator described
in the main paper. The actor maps the current observation $\mathbf{o}_t$ and
latent variable $\mathbf{z}_t$ to joint target positions $\mathbf{a}_t$. The
VAE combines an MLP for proprioceptive history $\mathbf{o}^{H}_t$ with a CNN
for exteroceptive observation $\mathbf{o}^{\text{extero}}_t$. Its MLP decoder
reconstructs the next proprioceptive state conditioned on $\mathbf{a}_t$, and
its CNN decoder reconstructs the exteroceptive observation. The detailed
hyperparameters are listed in \cref{tab:network_params}. All networks use ELU
activations, for which the MM propagation rule is derived in
\cref{sec:elu_derivation}.

\begin{table}[tb]
  \centering
  \setlength{\tabcolsep}{1mm}
  \begin{tabular}{@{}llr@{}}
      \toprule
      Module & Parameter & Value \\
      \midrule
      \multirow{2}{*}{Actor/Critic} & Units & $[512,256,128]$ \\
                                     & Activation & ELU \\
      \midrule
      \multirow{2}{*}{Proprio. Encoder (MLP)} & Units & $[256,256]$ \\
                                           & Latent Dim & $10$ \\
      \midrule
      \multirow{5}{*}{Extero. Encoder (CNN)}  & Channels & $[4, 8]$ \\
                                           & Kernels & $[3, 3]$ \\
                                           & Strides & $[2, 2]$ \\
                                           & FC Dim & $128$ \\
                                           & Latent Dim & $100$ \\
      \midrule
      \multirow{2}{*}{Proprio. Decoder (MLP)} & Units & $[256,256]$ \\
                                           & Recon. Dim & $96$ \\
      \midrule
      \multirow{4}{*}{Extero. Decoder (CNN)}  & FC Dim & $128$ \\
                                           & Channels & $[8, 4]$ \\
                                           & Kernels & $[3, 3]$ \\
                                           & Strides & $[2, 2]$ \\
      \midrule
      $\beta$-VAE & $\beta$ & $0.1$ \\
      \bottomrule
  \end{tabular}
  \caption{Network architecture used by the VAE baseline, MC-only estimators, and all $P^3$ variants.}
  \label{tab:network_params}
\end{table}

\subsection{Baseline Details}
\label{sec:baseline_detail}

\paragraph{SimpleActorCritic}
SimpleActorCritic is the direct Actor--Critic baseline in the main paper. It
operates without an encoder and directly takes the current proprioception
$\mathbf{o}_t^{\text{proprio}}$ and exteroception
$\mathbf{o}_t^{\text{extero}}$ as input. Because providing the full observation
history $\mathbf{o}^{H}_t$ to this uncompressed baseline substantially slows
convergence, SimpleActorCritic is configured to process single-step
observations only.

\paragraph{Self-Predictive Representations (SPR)}
The SPR baseline cited in the main paper constitutes a dynamics modeling approach that cultivates predictive latent representations by mandating consistency between predicted latent states and encoded future observations over multiple time steps.
The loss function for SPR is rigorously defined as:
\begin{equation}
    L_{\text{SPR}} = \sum_{k=1}^K \| f_\theta^{(k)}(z_t, a_{t:t+k-1}) - \text{sg}(g_\phi(o_{t+k})) \|_2^2,
\end{equation}
where $\text{sg}(\cdot)$ signifies the stop-gradient operation. 
Here, $f_\theta$ functions as the online dynamics model, recursively computing $z_{t+1} = f_\theta(z_t, a_t)$. 
Distinctly, $g_\phi$ serves as the target dynamics model, with its parameters evolving via an exponential moving average (EMA) of the online model's parameters to stabilize the learning target.
The loss coefficient $0.5$ and prediction steps $3$ are selected based on hyperparameter search.
SPR takes $\mathbf{o}_t^{\text{proprio}}$ and exteroception $\mathbf{o}_t^{\text{extero}}$ as input.

\paragraph{AutoEncoder}
The AutoEncoder (AE) baseline removes latent-space stochasticity from the VAE
while keeping the encoder, decoder, actor, critic, and all other architectural
elements identical. It therefore isolates the optimization effect of a
deterministic representation from the probabilistic propagation introduced by
$P^3$.

\subsection{PPO Optimization and Training Protocol}
\label{sec:algorithmic_details}
We adopt the PPO-clip algorithm for policy optimization. 
Unless stated otherwise, all training sessions utilize an adaptive learning rate mechanism. 
This approach dynamically scales the learning rate $\alpha_t$ in response to the aggregate KL divergence, serving as a critical stabilizer for the training process.
\begin{equation}
    \alpha_{t+1} = \begin{cases} 
        \max\left(10^{-5}, \frac{\alpha_t}{1.5}\right) & \text{if } \bar{D}_{\text{KL}} > 2 D_{\text{KL}}^{\text{desired}} \\ 
        \min\left(10^{-2}, 1.5 \alpha_t\right) & \text{if } 0 < \bar{D}_{\text{KL}} < \frac{D_{\text{KL}}^{\text{desired}}}{2} \\ 
        \alpha_t & \text{otherwise} 
    \end{cases}
\end{equation} 
Without this adaptation, some algorithms suffer from abrupt collapse during the later stages of training. 
The optimizer is Adam.
The selected hyperparameters align with standard configurations employed in contemporary reinforcement learning research for robotic locomotion control. 
\begin{table}[ht]
  \centering
  \begin{tabular}{@{}lc@{}}
      \toprule
      Parameter & Value \\
      \midrule
      Discount Factor $\gamma$ & $0.99$ \\
      GAE Parameter $\lambda$ & $0.95$ \\
      Number of Epochs & $5$ \\
      Number of Mini-batches & $4$ \\
      Entropy Coefficient & $0.01$ \\
      Clip Range $\epsilon$ & $0.2$ \\
      Value Loss Coefficient & $1.0$ \\
      Learning Rate & Adaptive \\
      Desired KL for Adaptive LR & $0.02$ \\
      \bottomrule
  \end{tabular}
  \caption{Hyperparameters for PPO training.}
  \label{tab:ppo_hyperparameters}
\end{table}

The estimator names and schedules follow the main paper. \emph{MC-only}
($N$) trains from initialization with $N$ latent samples, and the VAE baseline
is its $N=1$ case. \emph{$P^3$-MM} denotes the checkpoint obtained after 7,000
epochs of deterministic MM training. The complete \emph{$P^3$} schedule starts
from that checkpoint and performs 1,000 additional epochs of MC-based Latent
Sample Fine-Tuning (LSFT) with $N=15$. Other methods are trained until their
convergence criterion is met or until the 15,000-epoch budget is exhausted.
A run is marked as converged only after reaching a sustained near-asymptotic
plateau with terminal curriculum difficulty of at least 4.5.

\subsection{Environment and Terrain Details}
\label{sec:environment_details}

\begin{figure}[!t]
  \centering
  \includegraphics[width=0.6\columnwidth]{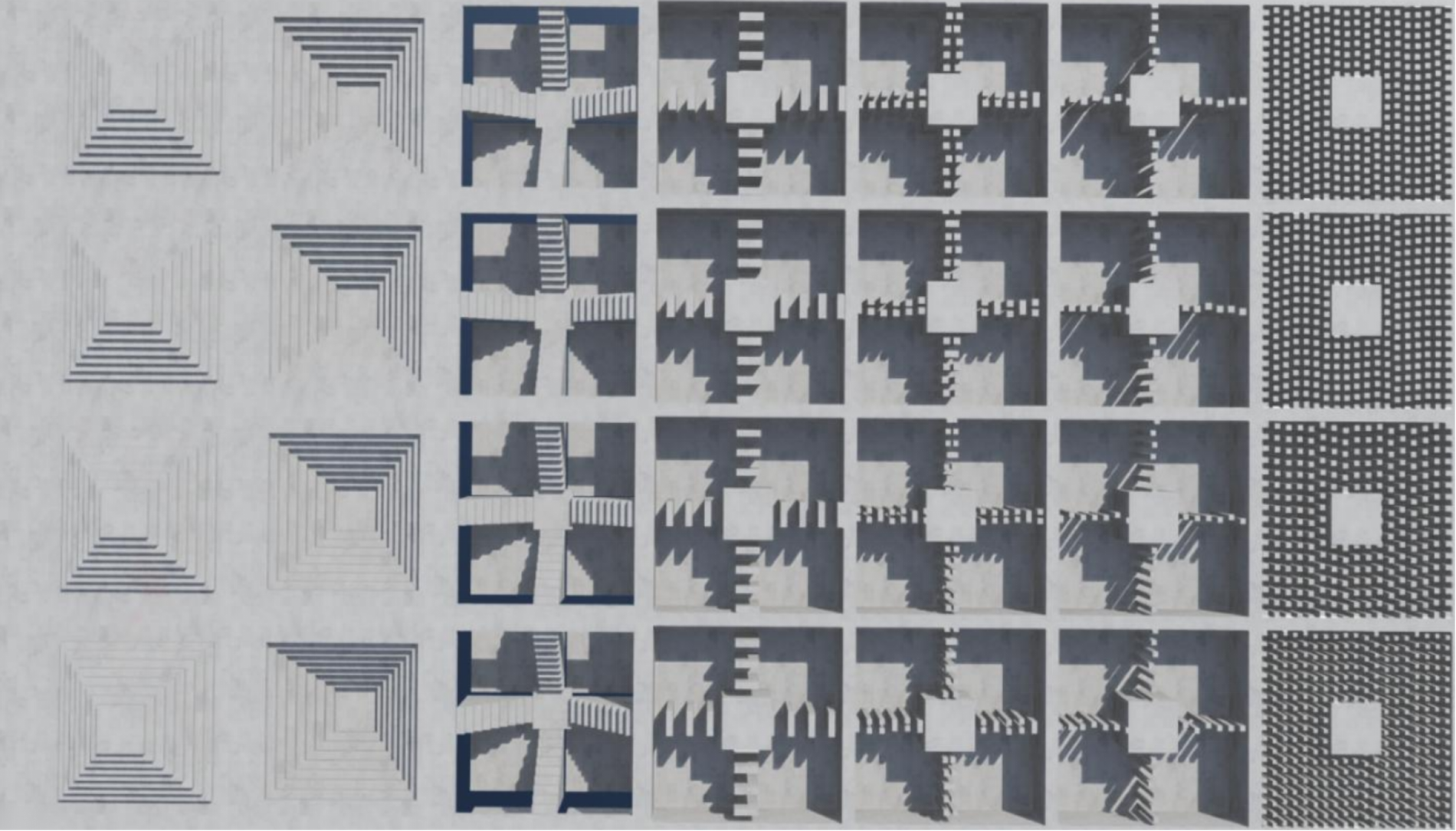}
  \caption{
    Terrain design. Terrain difficulty increases from top to bottom.
    From left to right, the terrains are pyramid stairs, inverted pyramid
    stairs, stairs, gaps, dual- and single-row stepping stones, and sparse
    stones.
  }
  \label{terrain_pic}
\end{figure}

We use 4096 parallel environments and collect 24 simulation steps from each
environment per PPO epoch.
The simulation frequency is 200 Hz, while the policy is trained and deployed at 50 Hz.
\Cref{terrain_pic} shows the terrains used in our experiments.

To improve the robustness of the policy and facilitate successful sim-to-real transfer, we extensively randomize the physical properties and observations during training. The randomization ranges are summarized in \cref{tab:domain_randomization}.

\begin{table}[tb]
  \centering
  \setlength{\tabcolsep}{1mm}
  \begin{tabular}{@{}p{0.47\columnwidth}p{0.43\columnwidth}@{}}
      \toprule
      Parameter & Range / Value \\
      \midrule
      \multicolumn{2}{@{}l}{\textit{Dynamics}} \\
      Friction & $[0.3,1.0]$ \\
      Restitution & $[0.0,0.1]$ \\
      Added Mass (Torso) & $[-1.0,3.0]$ kg \\
      CoM Displacement (Torso) & $x,y\in\pm0.05$ m; $z\in\pm0.01$ m \\
      Push Velocity (Interval 5--10 s) & $x,y\in[-0.5,0.5]$ m/s \\
      \midrule
      \multicolumn{2}{@{}l}{\textit{Observation Noise (Uniform)}} \\
      Base Angular Velocity & $\pm0.05$ rad/s \\
      Projected Gravity & $\pm0.05$ \\
      Joint Position & $\pm0.01$ rad \\
      Joint Velocity & $\pm0.1$ rad/s \\
      Height Scanner & $\pm0.01$ m \\
      \bottomrule
  \end{tabular}
  \caption{Domain randomization settings.}
  \label{tab:domain_randomization}
\end{table}

\section{Additional Experimental Results}

\subsection{Action-Distribution Diagnostics}
\label{subsec:action_distribution}
We select a model at epoch 1,000, well before the $P^3$-MM checkpoint reaches
its plateau at epoch 7,000 and within the common 15,000-epoch training budget.
This early checkpoint lets us test whether the policy already exhibits the
distributional properties required by latent-space sampling and the local
linearity assumption used in the theoretical analysis.
Notably, we observed that models from later training stages demonstrate even stronger linearity and an action mean distribution that more closely approximates a Gaussian. 

\paragraph{Shapiro-Wilk Test}
To quantify the marginal Gaussianity of the action-mean distribution, we
perform a Shapiro--Wilk test separately on each action dimension using action
means sampled from 4096 environments over 500 steps in Isaac Sim.
The results, averaged across action dimensions and shown in Table~\ref{tab:shapiro_wilk}, yield a Shapiro-Wilk $W$ statistic of 0.9995 and a p-value of 0.422.
Furthermore, with a significance level of $\alpha=0.05$, the pass rate is 85.39\% for action dimensions, which strongly supports the hypothesis that the distribution is approximately Gaussian.

\begin{table}[tb]
  \centering
  \setlength{\tabcolsep}{1mm}
  \begin{tabular}{@{}lccc@{}}
    \toprule
    Metric & W Statistic & p-value & \shortstack{Pass Rate\\($\alpha=0.05$)} \\
    \midrule
    Value & 0.9995 & 0.422 & 85.39\% \\
    \bottomrule
  \end{tabular}
  \caption{Shapiro--Wilk test results for the action-mean distribution.}
  \label{tab:shapiro_wilk}
\end{table}

\begin{figure}[tb]
  \centering
  \includegraphics[width=\linewidth]{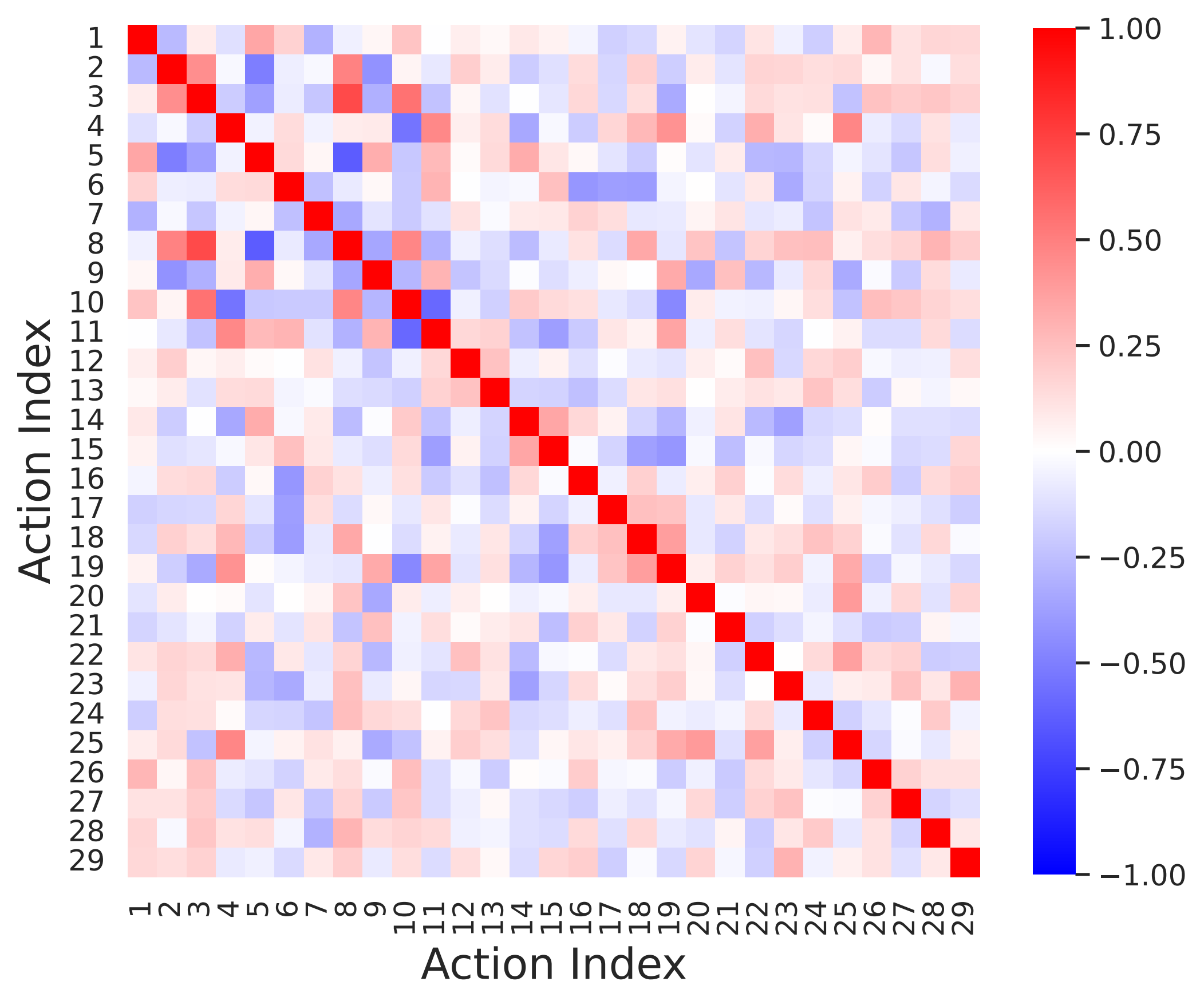}
  \caption{
    Correlation matrix of action dimensions 1--29 conditioned on latent samples at randomly selected states. Red and blue denote positive and negative correlation, respectively, over the range $[-1,1]$; the reported statistics are averaged over 4096 environments and 500 steps.
  }
  \label{fig:action_correlation}
\end{figure}

\paragraph{Cross-Dimensional Correlation Test}
We observe that the actor's output exhibits distinct correlation patterns across different time steps. 
To visualize these correlations, we present the covariance matrix of the action dimensions for a randomly selected time step in Figure~\ref{fig:action_correlation}.
The matrix reveals strong linear correlations among the action dimensions, explaining why the diagonal moment-matching (MM) approximation underestimates variance in our setting.

\paragraph{Latent Linearity Test}
To empirically validate the first-order Taylor approximation used in our gradient variance analysis (\cref{sec:variance_expansion}), we assess the linearity of the actor network's response to latent perturbations. 
Specifically, for each action dimension, we randomly sample 100 directions in the latent space and evaluate the actor's output along 50 uniformly spaced points in $2\sigma$ range in each direction. 
We then compute the coefficient of determination ($R^2$) of a linear regression fit to quantify the linearity of the mapping. 

\begin{figure}[tb]
  \centering
  \includegraphics[width=\linewidth]{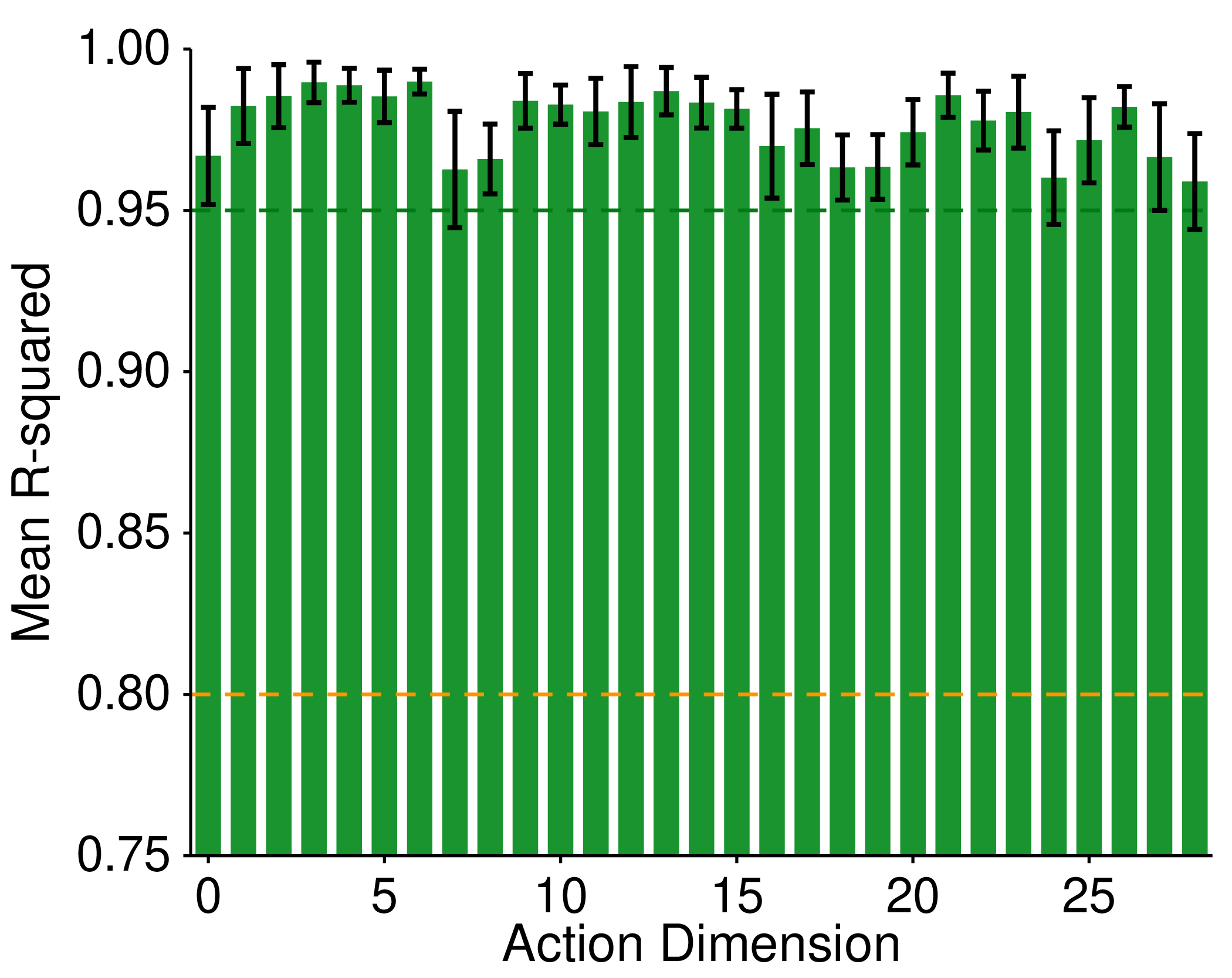}
  \caption{
    Latent linearity test across all action dimensions, showing the $R^2$ of a linear regression fit to the actor network's output along random directions in the latent space. The green and orange dashed lines mark thresholds of 0.95 and 0.80, respectively.
  }
  \label{fig:latent_linearity}
\end{figure}

As shown in Figure~\ref{fig:latent_linearity}, the overall average $R^2$ across all action dimensions is 0.9706, indicating that the actor network exhibits strong linear behavior with respect to latent variations.
This result directly supports the validity of the first-order Taylor expansion employed in our theoretical analysis. 

\subsection{Computational Cost}
\label{subsec:computational_cost}
We report the computational resource consumption of the estimators and
baselines in \cref{tab:computational_cost}. Memory usage is measured in MB.
The time entries report only the policy-learning time per PPO epoch in seconds;
they exclude environment interaction and data collection. In end-to-end
training, wall-clock time is dominated by data collection rather than policy
learning, so these values are provided only as a reference for relative
policy-learning overhead. The MC-only rows train with the indicated sample
count from initialization; $P^3$-MM uses the deterministic MM estimator before
LSFT.

\begin{table}[ht]
  \centering
  \setlength{\tabcolsep}{1mm}
  \begin{tabular}{@{}lrrrr@{}}
    \toprule
    \multicolumn{5}{c}{Marginalized-Policy Estimators} \\
    \cmidrule(lr){1-5}
    Metric & \shortstack{MC-only\\($N{=}5$)} & \shortstack{MC-only\\($N{=}15$)} & \shortstack{MC-only\\($N{=}50$)} & $P^3$-MM \\
    \midrule
    Memory (MB) & 14875 & 18457 & 31851 & 15138 \\
    Time (s) & 0.86 & 0.98 & 1.22 & 0.96 \\
    \bottomrule
  \end{tabular}

  \begin{tabular}{@{}lrrrr@{}}
    \toprule
    \multicolumn{5}{c}{Other Methods} \\
    \cmidrule(lr){1-5}
    Metric & \shortstack{SimpleActor\\Critic} & SPR & AE & VAE \\
    \midrule
    Memory (MB) & 13055 & 17541 & 13108 & 13438 \\
    Time (s) & 0.20 & 0.60 & 0.70 & 0.85 \\
    \bottomrule
  \end{tabular}
  \caption{Memory usage and policy-learning time per PPO epoch for MC-only estimators, $P^3$-MM, and other baselines.}
  \label{tab:computational_cost}
\end{table}

\FloatBarrier

\end{document}